\pdfoutput=1

\documentclass[11pt]{article}

\usepackage[]{acl}

\usepackage{times}
\usepackage{latexsym}

\usepackage[T1]{fontenc}

\usepackage[utf8]{inputenc}

\usepackage{microtype}

\usepackage{rotating}
\usepackage{amsmath,amsfonts,bm}

\def\eqref#1{equation~\ref{#1}}

\def\1{\bm{1}}

\DeclareMathAlphabet{\mathsfit}{\encodingdefault}{\sfdefault}{m}{sl}
\SetMathAlphabet{\mathsfit}{bold}{\encodingdefault}{\sfdefault}{bx}{n}

\DeclareMathOperator*{\argmax}{arg\,max}

\usepackage{booktabs}
\usepackage{xspace}
\usepackage{tabularx}
\usepackage{xhfill}
\usepackage{xcolor}
\usepackage{multirow}
\usepackage{mdframed}
\usepackage{tikz}
\usepackage{graphicx}
\usepackage{bbm}

\usepackage{pgfplots}
\pgfplotsset{compat=1.15}

\usepackage{mathtools}
\usepackage[colorinlistoftodos,prependcaption,textsize=tiny]{todonotes}
\usepackage{xargs}

\usepackage{amssymb}
\usepackage{pifont}

\usepackage{wrapfig}

\usepackage{ntheorem}
\theoremseparator{:}
\newtheorem{hyp}{Hypothesis}

\definecolor{forestgreen}{HTML}{009B55}
\definecolor{sepia}{HTML}{671800}
\definecolor{midnightblue}{HTML}{006795}
\definecolor{orangered}{HTML}{E24C00}
\definecolor{bblue}{HTML}{4F81BD}
\definecolor{rred}{HTML}{C0504D}
\definecolor{ggreen}{HTML}{9BBB59}
\definecolor{ppurple}{HTML}{9F4C7C}
\definecolor{olivegreen}{HTML}{556B2F}

\newcommand\gold{\textsc{Gold}\xspace}

\newcommand\pegasus{\textsc{Pegasus}\xspace}
\newcommand\fame{Focus (\textsc{Pegasus})\xspace}
\newcommand\frost{\textsc{Frost}\xspace}

\newcommand\content{\textsc{[content]}\xspace}
\newcommand\summary{\textsc{[summary]}\xspace}

\newcommand\rouge{\textsc{rouge}\xspace}

\newcommand\ednascore{\textsc{Edna}\xspace}

\title{A Well-Composed Text is Half Done!\\Composition Sampling for Diverse Conditional Generation}

\author{
Shashi Narayan \\ Google Research \\ \texttt{\small shashinarayan@google.com} \And
Gon\c{c}alo Sim\~{o}es \\ Google Research \\ \texttt{\small gsimoes@google.com} \And
Yao Zhao \\ Google Brain \\ \texttt{\small yaozhaoyz@google.com} \AND 
Joshua Maynez \\ Google Research \\ \texttt{\small joshuahm@google.com} \And
Dipanjan Das \\ Google Research \\ \texttt{\small dipanjand@google.com} \And
Michael Collins \\ Google Research \\ \texttt{\small mjcollins@google.com} \And
Mirella Lapata \\ Google Research \\ \texttt{\small lapata@google.com}
}

\begin{document}

\maketitle
\begin{abstract}

We propose Composition Sampling, a simple but effective method to
generate diverse outputs for conditional generation of higher quality
compared to previous stochastic decoding strategies. It builds on
recently proposed plan-based neural generation models \cite{frost}
that are trained to first create a composition of the output and then
generate by conditioning on it and the input.  Our approach avoids
text degeneration by first sampling a composition in the form of an
entity chain and then using beam search to generate the best possible
text grounded to this entity chain.
Experiments on summarization (CNN/DailyMail and XSum) and question
generation (SQuAD), using existing and newly proposed automatic
metrics together with human-based evaluation, demonstrate that
Composition Sampling is currently the best available decoding strategy
for generating diverse meaningful outputs.

\end{abstract}

\section{Introduction}
\begin{figure*}[t!]
\small
\setlength\tabcolsep{0.1cm}
\begin{tabular}{p{15.8cm}}
    \toprule \textbf{Haman Written Summary:} \underline{Chelsea} star
    \underline{Eden Hazard} is set to make his 100th top-flight
    appearance. \underline{Santi Cazorla} should hit the same
    milestone when \underline{Arsenal} meet \underline{Burnley}. Both
    players have impressed since moving to the \underline{Premier
      League} in \underline{2012}. \underline{Hazard} has more goals
    this season but \underline{Cazorla} has \underline{one} more
    assist. \underline{Sportsmail}'s reporters choose the player who
    has excited them the most.  \\ \midrule
     \textbf{Beam Search:} Eden Hazard and Santi Cazorla are both set to make their 100th Premier League appearances this weekend. Both players have been hugely influential since they moved to London. Here, Sportsmail's reporters choose the player they most enjoy seeing in action. \\  \midrule
     \textbf{Nucleus Sampling:} Eden Hazard and Santi \textcolor{orangered}{Cazorlag} will each make their 100th Premier League appearance this weekend. \textcolor{orangered}{nightstandapplication.com.} Sportsmail's \textcolor{orangered}{hovercraft} reporters choose their \textcolor{orangered}{man of the match countermeasures}. \\
      \textbf{Nucleus Sampling:} By making their 100th Premier League appearances this weekend, Eden Hazard and Santi \textcolor{orangered}{Cazor halibut} will set new records. Here, \textcolor{orangered}{Anna Coren} and \textcolor{orangered}{Dominic King} select their favourites. \\ \midrule
     \textbf{Composition Sampling:} \textcolor{gray}{(Eden Hazard $|$ Santi Cazorla $|$ Chelsea $|$ Arsenal $|$ Premier League $|||$ London $|$ 2012 $|||$)} Eden Hazard and Santi Cazorla are set to make their 100th appearances for Chelsea and Arsenal respectively in the Premier League this weekend. Both players have been hugely influential since they moved to London in the summer of 2012. But who has been the most exciting import to watch? \\
    \textbf{Composition Sampling:} \textcolor{gray}{(Chelsea $|$ Eden
      Hazard $|$ Arsenal $|$ Santi Cazorla $|||$ Sportsmail $|||$
      London)} Ch elsea's Eden Hazard and Arsenal's Santi Cazorla will both make 100th appearances this weekend. Sportsmail's reporters pick the player they most enjoy seeing in action. Both players have been hugely influential since moving to London. \\
    \bottomrule
\end{tabular}
\caption{Human written summary, single-best predicted summary using
  beam search (beam size $8$), and diverse summaries with nucleus
  sampling ($p=0.95$) and  our composition sampling for a
  CNN/DailyMail article (shown in the Appendix, Figure~\ref{fig:cnndm-article1}). 
We highlight spans in \textcolor{orangered}{orange} that are not faithful to the input.}
\label{fig:intro-cnndm-predictions-article1} 
\end{figure*}


In many NLG tasks, it is important to be able to generate multiple
diverse outputs from a model.  Tasks like
summarization \cite{mani2001automatic,Nenkova:McKeown:2011} and
question generation \cite{nqg2017} exhibit one-to-many relationships;
there can be multiple semantically diverse summaries or questions for
the same source, and it may be useful for a model to be able to
generate multiple outputs. Yet, the primary focus of recent research
in NLG has been on improving the quality of single-best
outputs \cite{t5,bart,unilm_arxiv19,zhang2019pegasus,frost}, while
diversity remains an unsolved
problem \cite{hashimoto-etal-2019-unifying,zhang-etal-2021-trading}. This
is particularly challenging in conditional generation, where diversity
in the target sequence should not come at the cost of correctness or
faithfulness; for example, alternate summaries are not valuable if
they are unfaithful to the input
document(s) \cite{maynez-etal-2020-faithfulness,kryscinski-etal-2020-evaluating}. In
this work, we investigate decoding methods for generating semantically
diverse text which is also faithful to its input focusing on two
tasks, namely summarization and question generation.


Beam search \cite{li-etal-2016-deep,wiseman-etal-2017-challenges} has proven successful for single-best 
generation
\cite{rush-etal-2015-neural,barrault-etal-2020-findings,meister-etal-2020-beam},
but struggles to generate diverse output
\cite{diversebeamsearch}. Stochastic sampling strategies, such as
\mbox{top-$k$} sampling \cite{fan-etal-2018-hierarchical} and nucleus
sampling \cite{nucleus}, are better at generating diverse
sequences but are not suitable for conditional
generation as they degenerate,\footnote{\newcite{nucleus} use the term
  `degeneration' to describe automatically generated text that is
  generic, repetitive, and awkward for story continuation. These
  issues are less common in conditional generation. In our case,
  `degenerate' refers to text unfaithful or inconsistent to the input.}
producing output that is not faithful to the
source. Figure~\ref{fig:intro-cnndm-predictions-article1} exposes
degeneration in summary output using nucleus sampling.

To address these shortcomings, we propose {\em Composition Sampling},
a simple but effective hybrid decoding method for diverse \emph{and}
faithful conditional generation. It builds on recently proposed
generation models \cite{frost} that are trained to first \emph{plan} a
semantic \emph{composition} of the target and then generate the text
conditioned on the composition \emph{and} the input.  Composition
sampling first samples a composition in the form of an \emph{entity chain}
and then uses beam search to generate the best possible
sequence \emph{grounded} to the sampled entity chain. Unlike top-$k$
or nucleus sampling, it avoids degeneration by instilling diversity in
composition, rather than directly on the surface form.



Our  contributions can be summarized  as follows: (a)~we introduce
Composition Sampling, a simple yet effective decoding method for
diverse conditional generation, which combines planning with
stochastic sampling;
(b)~we propose several metrics to compute semantic diversity in
generated text; our metrics are complementary to lexical diversity
(e.g.,~Self-BLEU; \citealt{texygen,alihosseini-etal-2019-jointly}) and
assess whether a set of diverse outputs are contextually dissimilar
(\emph{Self-BERTscore}; \citealt{bertscore}) or non-entailing
(\emph{Self-Entailment}); and (c)~finally, we introduce,
\textsc{Edna}, a novel metric aiming to ``\textbf{E}valuate
\textbf{D}iversity a\textbf{N}d f\textbf{A}ithfulness'' for
summarization by quantifying whether summaries in a diverse set are
faithful to their input without entailing each other. 

Evaluation on two popular summarization tasks,
namely highlight generation (CNN/DailyMail; \citealt{hermann-nips15}) and
extreme summarization (XSum; \citealt{narayan-etal-2018-dont}), and
question generation (SQuAD;
\citealt{rajpurkar-etal-2016-squad,nqg2017}), shows that composition
sampling is most effective in generating 
diverse summaries or questions. When assessed by humans, composition sampled summaries are as faithful as the
best summaries produced with beam search. In comparison, nucleus sampled summaries can be as diverse but  far less faithful. Taken together our results demonstrate  that
Composition Sampling is currently the best available decoding strategy
for generating diverse and meaningful output.\footnote{Our checkpoints and spaCy annotation code are available at \url{https://github.com/google-research/language/tree/master/language/frost}.}

\section{Background}

Conditional generation tasks such as summarization \cite{see-etal-2017-get}, data-to-text generation \cite{wiseman-etal-2017-challenges}, and machine translation \cite{Bahdanau2015NeuralMT}, are typically modeled using attention-based encoder-decoder architectures  \cite{Bahdanau2015NeuralMT,gu-etal-2016-incorporating,transformer}. The encoder first encodes the input text~$d$ and then the decoder predicts the output~$s_{1:n}$ (e.g.,~the translation or  summary of~$d$) one token at a time as $p(s_{i}|s_1,\ldots,s_{i-1};d)$, where, $n$ is the output length and $s_i$ is the $i$th token in the output. Often these models benefit from large scale \mbox{task-agnostic}  pretraining \cite{mass_icml19,gpt,bart,rothe-etal-2020-leveraging,t5,zhang2019pegasus}. 

\begin{figure*}[t!]
  \begin{center}
    \includegraphics[width=1\textwidth]{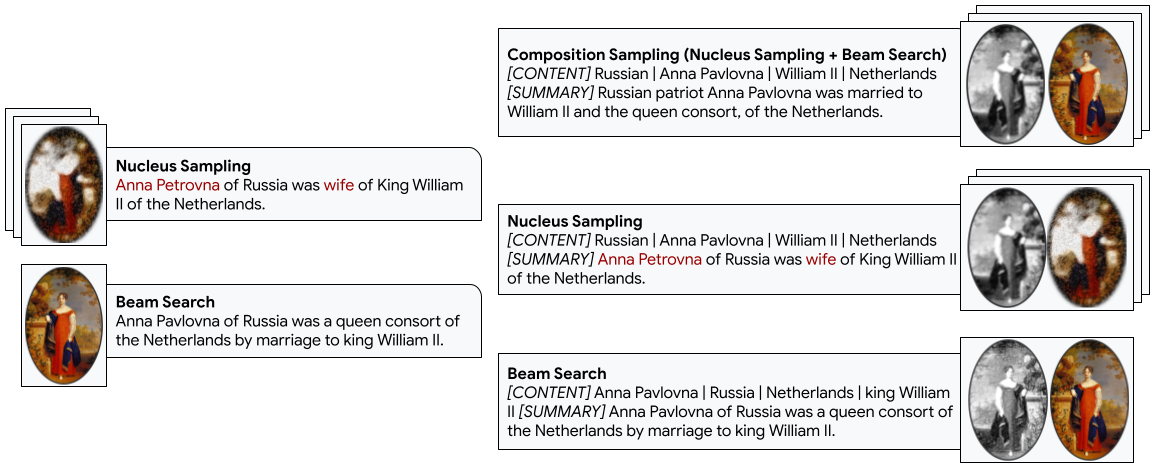} \\ 
  \end{center}
    \caption{Illustration of composition sampling and other decoding strategies with  vanilla and plan-based generation models. 
    The term `composition' is inspired from the quote ``{\em A Well-Composed Painting is Half Done}'' from French painter Pierre Bonnard. Images in black-and-white are early sketches or compositions of the painting in color. Nucleus or focus sampling often lead to hallucinations (highlight spans in \textcolor{red}{red}); corresponding color images are blurred to illustrate this. 
    (Credit: The image of ``Anna Pavlovna of Russia'' is taken from \href{https://en.wikipedia.org/wiki/Anna_Pavlovna_of_Russia\#/media/File:Anna_Pavlovna_of_Russia.jpg}{Wikipedia}.) }
    \label{fig:compositionsampling}
\end{figure*}



\paragraph{Plan-based Conditional Generation} 
\newcite{frost} develop a {\em \mbox{plan-based}}
approach for neural summarization; their decoder generates a
composition~$c_{1:m}$ of target summary~$s$ as
$p(c_{j}|c_1,\ldots,c_{j-1};d)$, and then the same decoder
produces~$s$ as~$p(s_{i}|s_1,\ldots,s_{i-1};c;d)$ conditioned on
input~$d$ and composition~$c_{1:m}$, with~$m$ being the composition
length. Specifically, they adopt entity chains as the
composition~$c$ of summary~$s$, under the assumption that entities in
the chain
ought to be observed in the output summary. During inference, the
model takes document~$d$ as input and generates~$c;s$, the
concatenation of composition and summary sequences, instead of
generating~$s$ directly; $c$ and $s$ are prefixed with special markers
``{\footnotesize \content}'' and ``{\footnotesize \summary}'',
respectively, as shown in Figure~\ref{fig:compositionsampling}. If $s$
consists of multiple sentences, markers``$|||$'' denote sentence
boundaries in composition~$c$.

The approach allows to directly manipulate the content of summaries
and their quality. For example, we might inspect the predicted chain
during inference and drop entities which are not present in the input
document, thereby controlling for hallucinations \cite{frost}. Outwith
summarization, similar constraints can be easily adapted to other
conditional generation tasks.

\paragraph{Maximization-Based Decoding}
In order to obtain the most likely output~$\hat{s}$ from
encoder-decoder models, we typically solve a maximization-based
objective: $\hat{x} = \argmax_x p(x|d)$,
where~$x$ is either the predicted output text~$s$ (for models without
planning) or the concatenation of the predicted composition and the
output text $c;s$ (for models with planning). It is standard practice
to use {\em beam
search} \cite{tillmann-ney-2003-word,li-etal-2016-deep,wiseman-etal-2017-challenges}
as solving the objective for the optimal sequence with neural sequence models is not tractable \cite{chen-etal-2018-recurrent}. 

\paragraph{Stochastic Sampling for Diverse Decoding}

Sampling-based strategies have been widely used to induce diversity in language models.
Temperature sampling uses a temperature to skew the distribution
towards high probability tokens at each decoding step
\cite{ackley:boltzmann,ficler-goldberg-2017-controlling,fan-etal-2018-hierarchical},
while \mbox{top-$k$} sampling truncates the distribution to~$k$ high
probability tokens
\cite{fan-etal-2018-hierarchical,holtzman-etal-2018-learning,gpt2}. Similarly
to \mbox{top-$k$} sampling, {\em nucleus sampling} \cite{nucleus} also
truncates the tail of the distribution but chooses~$k$ dynamically. At
each decoding step, it samples high-probable tokens from a nucleus~$N$
defined as the smallest subset of tokens from the vocabulary $V$ with
 cumulative probability $p'\geq p$, where $p$ is the pre-specified
mass of the
nucleus. 

\newcite{aralikatte-etal-2021-focus} introduce {\em focus sampling} to
promote diversity in summarization models. It constructs a subset $V_k
\subseteq V$ by sampling $k$ \mbox{source-relevant} and topical tokens
from the vocabulary distribution. Standard beam search decoding is
then used to generate a summary limited to~$V_k$. However, the authors
show that focus sampling is very sensitive to~$k$; increasing it
improves generation quality but at the cost of diversity.

\section{Composition Sampling}
\label{sec:compositionsampling}

{\em Composition Sampling} is a novel hybrid method which combines
stochastic sampling with maximization-based decoding, whilst leveraging
plan-based generation \cite{frost}. Specifically, we employ
nucleus sampling to obtain
 diverse 
compositions~$c_{\text{sample}}$ from $p(c|d)$ where $d$~is the input
text and $c$~are entity chains (prefixed with ``{\footnotesize \content}'' in
Figure~\ref{fig:compositionsampling}). We first employ nucleus
sampling to obtain diverse compositions from $p(c|d)$, where $d$ is
the input text. And then employ beam search to generate the
most-likely diverse output~$s$ (prefixed with ``{\footnotesize
  \summary}'' in Figure~\ref{fig:compositionsampling}), given
input~$d$ and composition~$c_{\text{sample}}$ as
$p(s|c_{\text{sample}};d)$. We will experimentally show that
composition sampling enables the 
generation of fluent, faithful and diverse texts for conditional
generation.

\paragraph{Why Entity Chains?} Unlike top-$k$ or nucleus sampling,
composition sampling avoids degeneration by introducing diversity in
composition, rather than directly on the surface form. For this to
effectively work, the choice of~$c$ needs to be well correlated with
an underlying notion of ``semantic composition'', which we want to
``diversify''; if $c_1$ and $c_2$ are two semantic compositions for
input $d$ such that $c_1 \neq c_2$, then two summaries $s_1 = \argmax_s
p(s|c_1;d)$ and $s_2 = \argmax_s p(s|c_2;d)$ are bound to be
diverse. In our work, we have chosen entity chains to model semantic
compositions; entity chains have been widely studied to model
entity-level lexical cohesion \cite{barzilay-elhadad-1997-using} and
coherence \cite{Halliday76a,azzam-etal-1999-using} in text. Also,
entity chains are unique to $d$, and thus can be easily distinguished
from compositions for other inputs. Moreover, entity chains provide a
very effective knob for content control in abstractive generation,
e.g., compositions can be constrained to entities only present in the input
document, thereby avoiding hallucinations and entity degeneration.


\begin{figure}[t]
\hspace*{-.25cm}
\begin{tikzpicture}

\definecolor{coral}{RGB}{255,127,80}
\definecolor{darkgray176}{RGB}{176,176,176}
\definecolor{lightgray204}{RGB}{204,204,204}
\definecolor{slategray}{RGB}{112,128,144}
\definecolor{teal}{RGB}{0,128,128}
\definecolor{whitesmoke}{RGB}{245,245,245}

\begin{axis}[
axis background/.style={fill=whitesmoke},
legend cell align={left},
legend style={fill opacity=0.8, draw opacity=1, text opacity=1,
  at={(0.6,1.18)}, draw=lightgray204, font=\tiny},
tick align=outside,
tick pos=left,
x grid style={darkgray176},
xmin=-3.35, xmax=70.35,
xtick style={color=black},
xtick={0,2,3,18,19,20,21,28,30,42,43,45,49,56,57,58,60,66,67},
xticklabel style={font=\tiny, rotate=90},
xticklabels={
  \_Chelsea,
  \_Eden,
\raisebox{-4ex}{\_Hazard},
 \raisebox{2ex}{\_Santi},
  \_Ca,
  \raisebox{-2ex}{zor},
\raisebox{-4ex}{la},
  \_Arsenal,
  \_Burnley,
  \raisebox{2ex}{\_Premier},
  \_League,
  \_2012,
  \_Hazard,
 \raisebox{2ex}{\_Ca},
  zor,
  la,
  \_one,
  \raisebox{2ex}{\_Sports},
  mail
},
y grid style={darkgray176},
ylabel={\tiny{Probability}},
ylabel style={at={(-0.08,0.5)}},
ymin=-0.0483847200755754, ymax=1.01791010230062,
ytick style={color=black},
tick label style={font=\tiny}  
  ]
\addplot [coral, dashed, mark=*, mark size=2, mark options={solid}]
table {%
0 0.0783894991507623
2 0.910721387501833
3 0.942324739251678
18 0.296310970533583
19 0.926034099962947
20 0.959748229360039
21 0.919216796187279
28 0.846395080634322
30 0.00221880525808118
42 0.598788998011347
43 0.910756132185516
45 0.885934844517921
49 0.0550409789147571
56 0.695699266769139
57 0.954385949616006
58 0.91057722261065
60 8.32263960697368e-05
66 0.143455628594203
67 0.935069987082599
};
\addlegendentry{Generate (Summary)}
\addplot [slategray, dash pattern=on 1pt off 3pt on 3pt off 3pt, mark=asterisk, mark size=2, mark options={solid}]
table {%
0 0.0900521078574531
2 0.401847305332897
3 0.95443874770712
18 0.0864469800821757
19 0.949828706112975
20 0.961195477679813
21 0.954721866678355
28 0.546893441958744
30 9.50317986252231e-05
42 0.0468281528639202
43 0.924762380885167
45 0.000197130029251411
49 0.139591078261183
56 0.350256862783432
57 0.965079883731889
58 0.938882838677528
60 0.0231704319052875
66 0.27611003919122
67 0.968075740933993
};
\addlegendentry{Plan-Generate (Entity Chain)}
\addplot [teal, dotted, mark=asterisk, mark size=2, mark options={solid}]
table {%
0 0.923163952724138
2 0.945883895019537
3 0.952979858870361
18 0.906497986472861
19 0.944388548058558
20 0.951371681985181
21 0.943288608142515
28 0.906657072697345
30 0.956564661682943
42 0.937578276209112
43 0.923156032011403
45 0.894186243001799
49 0.754641663711771
56 0.924960837688527
57 0.947462917494886
58 0.934325668547686
60 0.136157333217743
66 0.29689198298061
67 0.969442155828972
};
\addlegendentry{Plan-Generate (Summary)}
\end{axis}

\end{tikzpicture}
\caption{Probabilities of generating underlined entities in human
      written reference summary from
      Figure~\ref{fig:intro-cnndm-predictions-article1} (input article
      shown in Figure~\ref{fig:cnndm-article1}): when the summary is
      generated directly (Generate, Summary), when the entity chain
      ``{\em Chelsea $|$ Eden Hazard $|||$ Santi Cazorla $|$ Arsenal
        $|$ Burnley $|||$ Premier League $|$ 2012 $|||$ Hazard $|$
        Cazorla $|$ one $|||$ Sportsmail}'' is predicted first during planning (Plan-Generate, Entity Chain), and when the entities are predicted in the summary after planning (Plan-Generate,
      Summary). All probabilities were computed with \pegasus
      \citep{zhang2019pegasus} finetuned models. The symbol `\_' denotes start of token.
    }
    \label{fig:entprob-target}
    \end{figure}
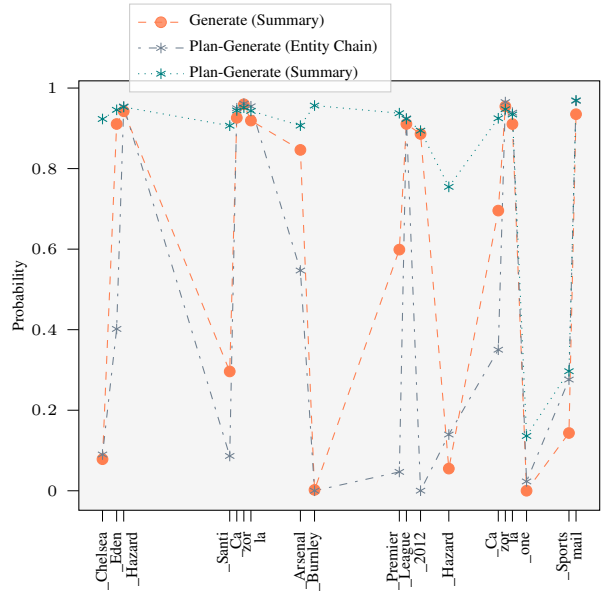
\begin{hyp} \label{hyp:entitychain}
If the semantic composition $c$ of the output text $s$ corresponds to
entity chains, then learning $p(c|d)$ is much easier than learning
$p(s|d)$; $d$ is the input. Hence, we can sample from $p(c|d)$ with
higher confidence than sampling directly from $p(s|d)$, and then
compute $\argmax_s p(s|c;d)$.
\end{hyp}

We demonstrate the effectiveness of entity chains as a choice for $c$
using the summarization example in
Figure~\ref{fig:entprob-target}. The negative log likelihood of
generating the summary~$s$ from scratch without planning ($- \log
p(s|d)$) is $121.18$, while the negative log likelihood of
generating composition~$c$ with planning ($- \log p(c|d)$) is $46.95$;
hence, the model is much more confident when sampling from $p(c|d)$
than directly from $p(s|d)$.


\paragraph{Why Grounded Generation?} 
The generation of~$s$ is inherently grounded to its entity
composition~$c$; following \newcite{frost}, the entity chains are
extracted from their targets during training. Hence, once the hard
part of planning the composition is done, the model is less perplexed
during generation of the output.

In Figure~\ref{fig:entprob-target}, the plan-based model is more
confident in predicting entities than its counterpart without
planning; perplexities of predicting entities in the summary with and
without planning are $0.24$ and $1.36$, respectively, and perplexities
of generating the whole summary with and without planning are $1.15$
and $1.48$, respectively. In fact, despite the increased length of the
target in the plan-based model (i.e.,~$c_{1:m};s_{1:n}$ instead of
$s_{1:n}$), we find that the perplexity of predicting the longer
sequence ($c_{1:m};s_{1:n}$) is lower than predicting just the summary 
without any planning, due to grounding ($1.16$ vs $1.48$). Overall,
$p(c;s|d)$, the plan-based approach, learns a more confident
distribution at each decoding step compared to no planing,
i.e.,~$p(s|d)$. For the example in Figure~\ref{fig:entprob-target},
the average cumulative probabilities for the top $15$ tokens in the
vocabulary distribution at each decoding step are $0.283$ for $p(s|d)$
and $0.433$ for $p(c;s|d)$.

In the following we assess composition sampling for its ability to
 generate semantically diverse output for two tasks, namely
 summarization (Section~\ref{sec:singledocsum}) and question
 generation (Section~\ref{sec:qgen}).

\section{Single Document Summarization}
\label{sec:singledocsum}

\subsection{Datasets and Models}


We evaluate our decoding strategy on two popular single document
summarization datasets: CNN/DailyMail highlight generation
\cite{hermann-nips15} and XSum extreme summarization
\cite{narayan-etal-2018-dont}, using the original train/validation/test splits. 
Inputs and outputs were truncated to 512 and 128 for XSum, and, 1,024
and 256 for CNN/DailyMail.\footnote{We also experimented with
  MultiNews \cite{fabbri-etal-2019-multi}, a multi-document
  summarization dataset. Results can be found in the Appendix (Table~\ref{table:multidoc-results}).}


We conduct experiments with state-of-the-art pretrained models for
summarization, namely \pegasus \cite{zhang2019pegasus} and \frost \cite{frost}. 
Our \pegasus finetuned model generates summaries directly, whereas \frost  generates    the  entity  chain  followed  by  the summary. 
In both cases we use large transformer architectures
\cite{transformer} with $L=16$, $H=1,024$, $F=4,096$, $A=16$ (568M
parameters), where~$L$ denotes the number of layers for encoder and
decoder Transformer blocks, $H$~is the hidden size, $F$~the
feed-forward layer size, and $A$~the number of self-attention
heads. Since this paper is proposing a decoding strategy, there is no
need to train new summarization models. We use the publicly available
\pegasus and \frost checkpoints.
Training details and model hyperparameters can be found in
\citet{zhang2019pegasus} and \citet{frost}. 

All models are decoded with a beam size of~$8$ and a length-penalty
of~$0.8$. For nucleus sampling and composition sampling, we use nucleus
probability $p$ set to~$0.95$.\footnote{Results with
  different temperatures and nucleus probabilities for random sampling, nucleus sampling, and
  composition sampling are in  Figure~\ref{fig:score_diff_weights}.} For focus sampling
\cite{aralikatte-etal-2021-focus}, we use $k=10,000$.


\subsection{Evaluation Metrics}

We assess our decoding strategy for likelihood, fluency, relevance,
faithfulness, and diversity, using both automatic and human
evaluation. \frost models predict a  plan in the form of an
entity chain, followed by a summary. All evaluations, except
likelihood, are done on the summary, while the predicted entity chains are
stripped out. For each \emph{diverse} decoding strategy, we sample 5 times
for each test document and report the average. 

\paragraph{Sequence Likelihood}
We report the perplexity of the generated sequence (i.e.,~entity
chains concatenated with their summaries for planning models and
summaries only for the others) using various decoding strategies.

\paragraph{Lexical Fluency and Relevance} 
We report {\em \rouge}-L F1 scores \cite{rouge} against reference
summaries.\footnote{
We lowercased
  candidate and reference summaries and used \texttt{pyrouge} with
  parameters ``-a -c 95 -m -n 4 -w 1.2.''}

\paragraph{Semantic Relevance} 
We report {\em BERTScore} \cite{bertscore} which computes the
contextual similarity between a candidate and its reference.

\paragraph{Faithfulness} 
We follow \newcite{maynez-etal-2020-faithfulness} and report on textual entailment \cite{pasunuru-bansal-2018-multi,falke-etal-2019-ranking,kryscinski-etal-2020-evaluating}. In particular, we report the probability of a summary entailing ({\em Entailment}) its input document using a classifier trained by fine-tuning an uncased BERT-Large pretrained model \cite{bert} on the Multi-NLI dataset \cite{williams-etal-2018-broad}. 

We further assess faithfulness by humans. Our annotators, proficient
in English, were tasked to read a document and then grade its summary
on a scale of 1--4 ({\em entirely unfaithful}, {\em somewhat
unfaithful}, {\em somewhat faithful}, and {\em entirely faithful}); a
summary is ``entirely faithful'' if its content is fully supported or
can be inferred from the document. We collected~3 ratings for each
(document, summary) pair; we report average system ratings (across
documents). With summaries deemed ``somewhat unfaithful'' and
``somewhat faithful'', annotators were asked to also specify what was
faithful or unfaithful, to improve agreement.

\begin{table*}[t!]
  \begin{center}{\small
  \begin{tabular}{l l | c c c | c c c } 
    \toprule
     &   Model & \multicolumn{3}{c}{XSum} & \multicolumn{3}{c}{CNN/DailyMail} \\
    & & R1 &  R2 &  RL & R1 & R2 & RL \\
    \midrule
&    GSum \citep{dou2020gsum} & 45.40& 21.89 & 36.67 & \textbf{45.94}& 22.32& 42.48 \\
&    CTRLsum  \citep{he2020ctrlsum} & ---& ---& --- & 45.65& \textbf{22.35}& \textbf{42.50} \\
&    \textsc{Fame}  \citep{aralikatte-etal-2021-focus} & 45.31& 22.75& 37.46 & 42.95 & 20.79 &39.90 \\
&    \pegasus \citep{zhang2019pegasus} & 47.56& 24.87& 39.40 & 44.05& 21.69& 40.98 \\
&    \frost \citep{frost} & \textbf{47.80}& \textbf{25.06}& \textbf{39.76} & 45.11& 22.11& 42.01\\
    \raisebox{.3cm}[0pt]{\begin{sideways}Single\end{sideways}} &
     \frost$_{\hspace*{-.1cm}++}$ \citep{frost}& 44.94& 21.58& 37.20 & 45.08& 22.14& 41.99  \\

    \midrule
&    Focus (\textsc{Fame}) & 42.76& 19.89& 34.97 & ---& ---& --- \\
&    Nucleus (\pegasus) & 38.49& 16.57& 30.99 & 36.27& 15.10& 33.46  \\
&    Nucleus (\frost) & 40.26& 17.83& 32.49 & 38.49& 15.71& 35.49   \\
&    Composition (\frost) & \textbf{45.12}& \textbf{22.24}& \textbf{36.98} & 41.76& 18.94& 38.69 \\
    \raisebox{.3cm}[0pt]{\begin{sideways}Diverse\end{sideways}}& Composition (\frost$_{\hspace*{-.1cm}++}$) & 43.82& 20.35& 35.89 & \textbf{42.37}& \textbf{19.48}& \textbf{39.28} \\
    \bottomrule
  \end{tabular}}
  \end{center}
  \caption{Comparison of decoding strategies with \rouge:
  single-best  vs diverse decoding (we report
    averages over 5 samples). Best results in each block are
    bold-faced. See  Table~\ref{table:xsum-cnndm-rouge-appendix} in
    the Appendix for more comparisons.}
  \label{table:xsum-cnndm-rouge}
\end{table*}



\paragraph{Diversity}

We report the number of times (out of five samples), a decoding
technique is able to generate a completely new summary ({\em Unique}).
We also use {\em Self-BLEU}
\cite{texygen,alihosseini-etal-2019-jointly} to measure lexical
diversity in the generated summaries. We consider all pairs of
summaries out of 5 samples, and for each pair we compute the
BLEU score \cite{papineni-etal-2002-bleu} considering one summary as a
hypothesis and the other as a reference. We report the average BLEU
score as the Self-BLEU of the document. The lower the Self-BLEU for a
decoding strategy is, the better it is in generating a more diverse set
of summaries.


We propose two additional measures to capture semantic diversity in
summaries: {\em Self-Entailment} and {\em Self-BERTScore}. Similar to
Self-BLEU, we compute the Entailment score and BERTScore for each
possible pair of summaries, respectively and report the average. A
lower Self-Entailment value suggests that the generated summaries do
not entail each other. Analogously, a low Self-BERTScore value
indicates that the decoding technique is able to generate more
contextually dissimilar summaries.

We further assess diversity by humans. Our annotators, proficient in
English, again read two summaries (out of five samples) and then
graded the pair on a scale of 1--4 ({\em identical}, {\em somewhat
identical}, {\em somewhat diverse}, and {\em diverse}); the document
was not shown in this assessment.  Two summaries are ``identical'' if
they are semantically equivalent, while the same information may be
presented differently in the case of ``somewhat identical''. A
``somewhat diverse'' pair may introduce one or two new concepts or
topics in one summary. A ``diverse'' pair should introduce new
concepts or topics in each summary. We collected~3 ratings for each
pair and report their average. This assessment was only done with
single-sentence XSum summaries, in  future work we will explore how
to do this effectively for multi-sentence summaries.

\begin{table*}[t!]
  \begin{center}{\small 
  \begin{tabular}{@{}l l l| c | c c @{~}| c@{~}c @{~}| c@{~}c@{~}c@{~}c@{~}c@{~}| c@{}} 
    \toprule
    \multicolumn{3}{c|}{\multirow{2}{*}{{Models}}} & \multirow{2}{*}{{ppl}}  & \multicolumn{2}{c|}{{With Ref.}} & \multicolumn{2}{c|}{{Faithfulness}} & \multicolumn{5}{c|}{{Diversity}} & {Div+Faith} \\
    & & & & {RL} & {BSc} & {Ent} & {Human} & {Uniq} & {S-BL} & {S-Ent} &  {S-BSc} & {Human} & {\ednascore}\\
    \midrule 
    \multirow{9}{*}{\rotatebox[origin=c]{90}{{XSum}}} & \multirow{4}{*}{\rotatebox[origin=c]{90}{{Single}}} & 
    \textsc{Fame}  & --- & 34.23 & 0.70 & 0.24 & 2.19 &  --- & --- & --- & --- & --- & ---\\
    & & \pegasus & 0.51 & 40.69 & 0.76 & 0.40 & 2.52 &  --- & --- & --- & --- & --- & ---\\
    & & \frost & 0.31 & 40.90 & 0.75 & 0.37 & 2.63 & --- & --- & --- & --- & --- & ---\\
    & & \frost$_{\hspace*{-.1cm}++}$ & 0.71 & 33.75 & 0.70 & 0.44 & 2.78 & --- & --- & --- & --- & --- & --- \\
    \cline{2-14}
    & \multirow{5}{*}{\rotatebox[origin=c]{90}{{Diverse}}} & Focus (\textsc{Fame})  & --- & 29.19 & 0.66 & 0.23 & 1.88 & 2.6 & 89.51 & 0.62 & 0.91 & 1.84 & 0.09 \\
    & & Nucleus (\pegasus) & 1.47 & 31.10 & 0.68 &  0.24 & 2.00 & \textbf{5.0} &  \textbf{26.22} & \textbf{0.10} & \textbf{0.68} & \textbf{3.11} & 0.30 \\
    & & Nucleus (\frost) & 0.83 & 33.81 & 0.71 & 0.22 & 2.11 & \textbf{5.0} &  31.08 & \textbf{0.10} & 0.71 & 3.08 & 0.27 \\
    & & Composition (\frost) & \textbf{0.51} & \textbf{36.95} & \textbf{0.73} & 0.27 & 2.37 & 4.7 &  58.94 & 0.17 & 0.79 & 2.73 & 0.30 \\
    & & Composition (\frost$_{\hspace*{-.1cm}++}$) & 0.74 & 33.87 & 0.70 & \textbf{0.43} & \textbf{2.75} & 3.5 &  76.87 & 0.40 & 0.84 & 2.25 & \textbf{0.35} \\
    \midrule 
    \midrule 
    \multirow{7}{*}{\rotatebox[origin=c]{90}{{CNN/DM}}} & \multirow{3}{*}{\rotatebox[origin=c]{90}{{Single}}} & \pegasus & 0.35 & 36.09 & 0.65 & 0.70 & 3.78 & --- & --- & --- & --- & --- & --- \\
    & & \frost &  0.30 & 39.03 & 0.66 & 0.72 & 3.74 &  --- & --- & --- & --- & --- & ---\\
    & & \frost$_{\hspace*{-.1cm}++}$ & 0.37 & 38.87 & 0.66 & 0.79 & 3.94 & --- & --- & --- & --- & --- & ---\\
    \cline{2-14}
    & \multirow{4}{*}{\rotatebox[origin=c]{90}{{Diverse}}} & Nucleus (\pegasus) & 1.39 & 28.99 & 0.62 & 0.62 & 3.08 & \textbf{5.0} & \textbf{26.99} & 0.03 & \textbf{0.63} & --- & 0.70 \\
    & & Nucleus (\frost) & 1.04 &  31.58 & 0.63 & 0.56 & 3.08 & \textbf{5.0} & 29.60 & \textbf{0.03} & 0.64 & --- & 0.66 \\
    & & Composition (\frost) & 0.52 &  35.06 & \textbf{0.64} & 0.59 & 3.45 & \textbf{5.0} & 58.60 & 0.04 & 0.71 & --- & 0.66 \\
    & & Composition (\frost$_{\hspace*{-.1cm}++}$) & \textbf{0.46} &  \textbf{35.07} & 0.64 & \textbf{0.73} & \textbf{3.89} & 4.9 & 62.81 & 0.07 & 0.72 & --- & \textbf{0.78}\\
    \bottomrule
  \end{tabular}}
  \end{center}
  \caption{Likelihood, faithfulness and diversity results on 50
    documents sampled from XSum and CNN/DailyMail each. 
  We report on perplexity (ppl), Entailment (Ent), Uniqueness (Uniq), Self-BLEU (\mbox{S-BL)}, Self-Entailment (\mbox{S-Ent}), Self-BERTScore (S-BSc) and \ednascore scores, along with \rouge (RL) and BERTScore (BSc) for comparison. We also report on human judgements for faithfulness and diversity. 
Additional R1 and R2 numbers can be found in the Appendix (Table~\ref{table:xsum-cnndm-all-results-appendix}). 
Best results in the diverse block for each dataset are bold faced. Scores for single-best decoded summaries are also presented for comparison. 
Focus (\textsc{fame}) diverse predictions on  CNN/DailyMail are not available.
The lower the Self-* metric is, the better the decoding method in generating diverse outputs.
}
  \label{table:xsum-cnndm-all-results}
\end{table*}

\paragraph{Diversity and Faithfulness}

For summarization, diverse summaries are not meaningful if they are
not faithful to the input. We propose {\em \ednascore}, a novel measure for
``\textbf{E}valuating \textbf{D}iversity a\textbf{N}d
f\textbf{A}ithfulness'' in summaries.  \ednascore is the harmonic mean of
Entailment and ($1-$Self-Entailment); higher values of \ednascore imply
more faithful and diverse summaries. The reason \ednascore relies on
Self-Entailment to measure diversity is because the faithfulness
metric is also based on Entailment. This means that both
components  will be mapped to a score in a similar output
space (i.e.,~they both yield values between 0 and 1 obtained through
the same trained model), making it more likely to be
properly balanced when mixed.


\subsection{Results}

Table~\ref{table:xsum-cnndm-rouge} presents \rouge results on the XSum
and CNN/DailyMail test sets. The top block includes results for models
which employ maximization-based decoding.  GSum \citep{dou2020gsum} is
a state-of-the art system which decodes summaries guided by an an
extractive model at test time. CTRLsum \citep{he2020ctrlsum} controls
the summarization output trough keywords and automatically extracted
sentences.  \textsc{Fame} \citep{aralikatte-etal-2021-focus} uses a
focus attention mechanism to encourage the decoder to proactively
generate tokens that are similar or topical to the input document.  As
mentioned earlier \frost \citep{frost} first generates an entity chain
and then a summary while \frost$_{\hspace*{-.1cm}++}$ is a constrained
variant which restricts the predicted entities to those present in the
document. We also show results for a vanilla \pegasus model
\citep{zhang2019pegasus} finetuned on our datasets.

The bottom block focuses on diverse decoding (we report averages
across five samples). We show results with Focus sampling
\cite{aralikatte-etal-2021-focus} built on top of \textsc{Fame},
Nucleus sampling \cite{nucleus} with \pegasus and \frost, and our
Composition sampling.

Table~\ref{table:xsum-cnndm-all-results} presents more detailed
faithfulness and diversity results, on challenge sets consisting of 50
documents for each XSum and CNN/DailyMail summaries.  We construct
these challenge sets by randomly selecting documents whose reference
summaries have non-extractive entity chains in them; an entity chain
is extractive if all entities in it can be found in the input
document. \newcite{frost} have found that models struggle to generate
faithful summaries for documents with data-divergence issues
\cite{dhingra-etal-2019-handling}. The same challenge sets were used
for our human evaluations of faithfulness and diversity.


\paragraph{Composition Sampling is not as Performance Diminishing as
  Nucleus Sampling} Single-best decoding for \frost achieves 39.76
\rouge (RL) on XSum,; nucleus and composition sampling fare worse
showing an average drop of 7.27 and~2.78, respectively. Similarly, for
CNN/DailyMail, \rouge drops for nucleus sampling by an average of~6.51
points, compared to an average drop of 3.28 points for composition
sampling (with \frost). Nucleus sampling is even more damaging for
non-plan based models, such as \pegasus; we see an average drop of
8.59 and 7.30 \rouge points on XSum and CNN/DailyMail.  These gaps are
slightly larger for the challenging subsets in
Table~\ref{table:xsum-cnndm-all-results} which is expected due to the
highly abstractive nature of the reference summaries therein. 

On XSum, composition Sampling with \mbox{\frost$_{\hspace*{-.1cm}++}$}
performs slightly worse than with vanilla \frost in terms of \rouge. This is due to the extreme abstractive nature of the XSum reference summaries
\cite{maynez-etal-2020-faithfulness}; as a result, a model is required
to hallucinate factual content, that is not necessarily faithful to
the input (see examples of XSum summaries in the Appendix,
Figure~\ref{fig:xsum-predictions-all}). But
Composition(\mbox{\frost$_{\hspace*{-.1cm}++}$}) only keeps supported
entities in the sampled plans giving rise to summaries which diverge
from their reference. This is not the case with CNN/DailyMail which is
mostly extractive and we see that \rouge performance improves with
Composition(\mbox{\frost$_{\hspace*{-.1cm}++}$}) in
Table~\ref{table:xsum-cnndm-rouge}.



\paragraph{Composition Sampling is more Confident in its Predictions than Nucleus Sampling}
Perplexity for \frost predictions increases from $0.31$ to $0.83$ for
nucleus sampling, but only to $0.51$ for composition sampling, on
XSum. \pegasus shows an even larger increment in perplexity (from
$0.51$ to $1.47$) for nucleus sampling. Similar patterns are observed
for  CNN/DailyMail summaries.  

Composition(\frost$_{\hspace*{-.1cm}++}$) is more perplexed when
generating XSum summaries due to the reference summary divergence
issue discussed earlier; perplexity increases from $0.51$ to $0.74$
compared to Composition(\frost). Interestingly,
Composition(\frost$_{\hspace*{-.1cm}++}$) is almost as confident in
generating diverse summaries as single-best beam decoding (\frost;
perplexities of $0.71$ vs $0.74$ for XSum). Unsurprisingly,
Composition(\frost$_{\hspace*{-.1cm}++}$) is more confident in
generating CNN/DailyMail summaries than \frost ($0.46$ vs $0.52$) due
to their extractive nature.

\paragraph{Constrained Composition is Most Effective in Generating
  Meaningful Diverse Summaries}
It is no surprise that nucleus sampling is able to generate the most
diverse summaries on both XSum and CNN/DailyMail (achieving best
scores for \mbox{Self-BLEU}, Self-Entailment, Self-BERTScore, and diversity
assessed by humans); however these summaries perform poorly on
faithfulness measures. Composition(\frost$_{\hspace*{-.1cm}++}$) is
most effective in generating faithful summaries, as demonstrated
automatically (with best entailment scores on XSum and CNN/DailyMail)
and by humans (with highest ratings on XSum and CNN/DailyMail); these
summaries are also diverse achieving highest \ednascore scores on both
summarization datasets.

We also examined whether models differ in terms of faithfulness and
diversity as rated by our participants. We carried out pairwise
comparisons using one-way ANOVA with post-hoc Tukey HSD tests
(\mbox{$p < 0.01$}). The difference between \mbox{Nucleus(\pegasus)}
and \mbox{Nucleus(\frost)} is not significant. Nucleus(\pegasus) was
also not significantly more faithful than Focus(\textsc{Fame}) for XSum
summaries. All other pairwise differences were significant for both
faithfulness and diversity.

In sum, our results demonstrate that composition sampling is a better
alternative to nucleus or focus sampling for generating meaningful
diverse summaries. Figure~\ref{fig:intro-cnndm-predictions-article1}
presents summaries from different decoding strategies for a
CNN/DailyMail article. Other example predictions for XSum and
CNN/DailyMail articles can be found in the Appendix
(Figures~\ref{fig:xsum-predictions-all}--\ref{fig:composition-predictions-article2}).

\begin{table}[t!]
\begin{center}
  \begin{small}
  \begin{tabular}{l | r r } 
    \toprule
    {Metric} & \multicolumn{1}{c}{Faithfulness} & \multicolumn{1}{c}{Diversity} \\
    \midrule 
    \rouge-L & 0.197 & $-$0.164 \\
    BERTScore & 0.209 & $-$0.195\\ 
    Entailment & {0.588} & $-$0.067 \\ 
    1 - Self-BLEU  & $-$0.208 & {0.880} \\
    1 - Self-Entailment  & $-$0.187 & 0.771 \\
    1 - Self-BERTScore & $-$0.198 & 0.873 \\ 
    \ednascore & 0.482 & 0.174 \\
    \bottomrule
  \end{tabular}
  \end{small}
  \end{center}
  \caption{Different automatic metrics and their correlation against  human
    assessments using Spearman's rank coefficient.}
  \label{table:xsum-correlation-results}
\end{table}

\pgfplotstableread[row sep=\\,col sep=&]{
weights	& PegBeam & PegRandom & PegNucleus & FrostBeam & FrostConstBeam & FrostRandom & FrostNucleus & FrostComp & FrostCompConst \\
0.200 & 0.350 & 0.367 & 0.462 & 0.300 & 0.370 & 0.303 & 0.370 & 0.324 & 0.376\\
0.400 & 0.350 & 0.397 & 0.437 & 0.300 & 0.370 & 0.336 & 0.348 & 0.318 & 0.362\\
0.600 & 0.350 & 0.489 & 0.467 & 0.300 & 0.370 & 0.376 & 0.378 & 0.327 & 0.371\\
0.800 & 0.350 & 0.693 & 0.578 & 0.300 & 0.370 & 0.537 & 0.471 & 0.358 & 0.388\\
0.950 & 0.350 &  & 1.378 & 0.300 & 0.370 &  & 1.063 & 0.494 & 0.441\\
1.000 & 0.350 & 2.060 & 2.010 & 0.300 & 0.370 & 1.618 & 1.773 & 0.702 & 0.498\\
}\ppl
\pgfplotstableread[row sep=\\,col sep=&]{
weights	& PegBeam & PegRandom & PegNucleus & FrostBeam & FrostConstBeam & FrostRandom & FrostNucleus & FrostComp & FrostCompConst \\
0.200 & 0.700 & 0.713 & 0.736 & 0.723 & 0.794 & 0.743 & 0.661 & 0.665 & 0.736\\
0.400 & 0.700 & 0.751 & 0.683 & 0.723 & 0.794 & 0.711 & 0.720 & 0.691 & 0.755\\
0.600 & 0.700 & 0.796 & 0.649 & 0.723 & 0.794 & 0.704 & 0.690 & 0.706 & 0.771\\
0.800 & 0.700 & 0.788 & 0.678 & 0.723 & 0.794 & 0.671 & 0.677 & 0.650 & 0.777\\
0.950 & 0.700 &  & 0.590 & 0.723 & 0.794 &  & 0.532 & 0.623 & 0.764\\
1.000 & 0.700 & 0.515 & 0.487 & 0.723 & 0.794 & 0.391 & 0.385 & 0.504 & 0.722\\
}\entailment
\pgfplotstableread[row sep=\\,col sep=&]{
weights	& PegRandom & PegNucleus & FrostRandom & FrostNucleus & FrostComp & FrostCompConst \\
0.200 & 0.440 & 0.533 & 0.331 & 0.613 & 0.787 & 0.783\\
0.400 & 0.209 & 0.099 & 0.154 & 0.194 & 0.317 & 0.308\\
0.600 & 0.130 & 0.057 & 0.114 & 0.133 & 0.171 & 0.166\\
0.800 & 0.076 & 0.041 & 0.049 & 0.070 & 0.102 & 0.103\\
0.950 &  & 0.055 &  & 0.041 & 0.071 & 0.075\\
1.000 & 0.020 & 0.049 & 0.036 & 0.021 & 0.035 & 0.081\\
}\selfentailment
\pgfplotstableread[row sep=\\,col sep=&]{
weights	& PegRandom & PegNucleus & FrostRandom & FrostNucleus & FrostComp & FrostCompConst \\
0.200 & 0.627 & 0.571 & 0.704 & 0.488 & 0.323 & 0.335\\
0.400 & 0.770 & 0.777 & 0.773 & 0.761 & 0.687 & 0.722\\
0.600 & 0.831 & 0.769 & 0.785 & 0.768 & 0.763 & 0.801\\
0.800 & 0.851 & 0.794 & 0.787 & 0.784 & 0.754 & 0.833\\
0.950 &  & 0.726 &  & 0.684 & 0.746 & 0.837\\
1.000 & 0.675 & 0.644 & 0.556 & 0.553 & 0.662 & 0.809\\
}\edna
\begin{figure*}[t!]
\small
  \center{
\begin{tabular}{@{~}c@{~}c@{~}c@{~}c@{~}}
\hspace*{-.2cm}\begin{tikzpicture}[scale=0.54]
\begin{axis}[
        name=plot1,
        width=3.0in,
        height=2.0in,
        title={},
        xlabel={\small{Different Temperatures or Nucleus Probabilities}},
ylabel={\small{Perplexity}},
        xmin=0.18, xmax=1.02,
        ymin=0.2, ymax=1.8,
        xtick={0.2, 0.4, 0.6, 0.8, 0.9, 1.0},
        ytick={},
        xticklabels={\small{0.2}, \small{0.4}, \small{0.6}, \small{0.8}, \small{0.9}, \small{1.0}},
        legend pos=south west,
        legend style={at={(0.05,0.5)}},
        xmajorgrids=true,
        ymajorgrids=true,
        grid style=dashed,
        legend style={font=\small},
        ]
        \addplot [ppurple] table[x=weights,y=FrostBeam]{\ppl};
        \addplot [forestgreen] table[x=weights,y=FrostConstBeam]{\ppl};
        \addplot [sepia] table[x=weights,y=FrostRandom]{\ppl};
        \addplot [midnightblue] table[x=weights,y=FrostNucleus]{\ppl};
        \addplot [orangered] table[x=weights,y=FrostComp]{\ppl};
        \addplot [olivegreen] table[x=weights,y=FrostCompConst]{\ppl};
                         \legend{
         Beam(\frost), Beam(\frost$_{\hspace*{-.1cm}++}$), Random(\frost), Nucleus(\frost), Composition(\frost), Composition(\frost$_{\hspace*{-.1cm}++}$)}

      \end{axis}
\end{tikzpicture}
&
\begin{tikzpicture}[scale=0.54]

      \begin{axis}[
        name=plot2,
        at=(plot1.below south east), anchor=above north east,
        width=3.0in,
        height=2.0in,
        title={},
        xlabel={\small{Different Temperatures or Nucleus Probabilities}},
ylabel={\small{Entailment (Faithfulness)}},
        xmin=0.18, xmax=1.02,
        ymin=0.36, ymax=0.81,
        xtick={0.2, 0.4, 0.6, 0.8, 0.9, 1.0},
        ytick={},
        xticklabels={\small{0.2}, \small{0.4}, \small{0.6}, \small{0.8}, \small{0.9}, \small{1.0}},
        legend pos=outer north east,
        xmajorgrids=true,
        ymajorgrids=true,
        grid style=dashed,
        legend style={font=\small},
        legend pos=outer north east
        ]
        \addplot [ppurple] table[x=weights,y=FrostBeam]{\entailment};
        \addplot [forestgreen] table[x=weights,y=FrostConstBeam]{\entailment};
        \addplot [sepia] table[x=weights,y=FrostRandom]{\entailment};
        \addplot [midnightblue] table[x=weights,y=FrostNucleus]{\entailment};
        \addplot [orangered] table[x=weights,y=FrostComp]{\entailment};
        \addplot [olivegreen] table[x=weights,y=FrostCompConst]{\entailment};
      \end{axis}
      \end{tikzpicture} &
\begin{tikzpicture}[scale=0.54]
\begin{axis}[
        name=plot3,
        at=(plot2.below south east), anchor=above north east,
        width=3.0in,
        height=2.0in,
        title={},
        xlabel={\small{Different Temperatures or Nucleus Probabilities}},
ylabel={\small{Self-Entailment (Diversity)}},
        xmin=0.18, xmax=1.02,
        ymin=0, ymax=0.81,
        xtick={0.2, 0.4, 0.6, 0.8, 0.9, 1.0},
        ytick={},
        xticklabels={\small{0.2}, \small{0.4}, \small{0.6}, \small{0.8}, \small{0.9}, \small{1.0}},
         legend pos=south east,
        xmajorgrids=true,
        ymajorgrids=true,
        grid style=dashed,
        legend style={font=\small},
        ]
        \addplot [sepia] table[x=weights,y=FrostRandom]{\selfentailment};
        \addplot [midnightblue] table[x=weights,y=FrostNucleus]{\selfentailment};
        \addplot [orangered] table[x=weights,y=FrostComp]{\selfentailment};
        \addplot [olivegreen] table[x=weights,y=FrostCompConst]{\selfentailment};

\end{axis}
      \end{tikzpicture} &
      \begin{tikzpicture}[scale=0.54]
      \begin{axis}[
        name=plot4,
        at=(plot3.below south east), anchor=above north east,
        width=3.0in,
        height=2.0in,
        title={},
        xlabel={\small{Different Temperatures or Nucleus Probabilities}},
        ylabel={\small{\ednascore (Faithfulness+Diversity)}},
        xmin=0.18, xmax=1.02,
        ymin=0.3, ymax=0.85,
        xtick={0.2, 0.4, 0.6, 0.8, 0.9, 1.0},
        ytick={},
        xticklabels={\small{0.2}, \small{0.4}, \small{0.6}, \small{0.8}, \small{0.9}, \small{1.0}},
        legend pos=outer north east,
        xmajorgrids=true,
        ymajorgrids=true,
        grid style=dashed,
        legend style={font=\small},
        ]
        \addplot [sepia] table[x=weights,y=FrostRandom]{\edna};
        \addplot [midnightblue] table[x=weights,y=FrostNucleus]{\edna};
        \addplot [orangered] table[x=weights,y=FrostComp]{\edna};
        \addplot [olivegreen] table[x=weights,y=FrostCompConst]{\edna};
      \end{axis}
    \end{tikzpicture}\\
a. & b.&    c. & d. 
\end{tabular}
}
\vspace{-0.3cm}
  \caption{Perplexity, entailment, self-entailment and \ednascore scores on
 the CNN/DailyMail challenge set
 (Table~\ref{table:xsum-cnndm-all-results}) with varying temperatures
 (for random sampling) and nucleus Probabilities (for nucleus and
 composition sampling). For each diverse decoding strategy, we sample
 5 times per document and report the average. \label{fig:score_diff_weights}}
\end{figure*}
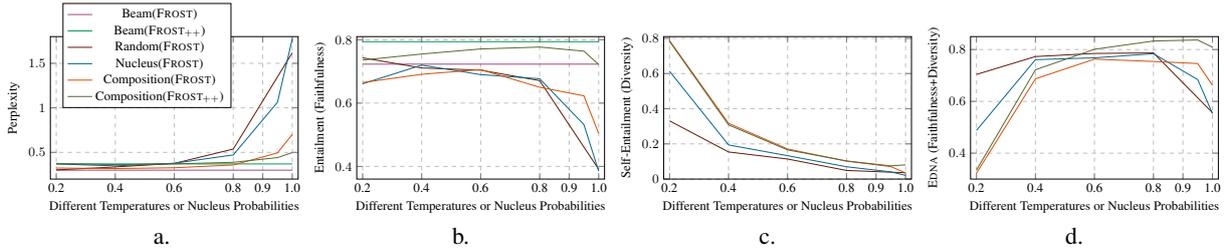

\paragraph{Faithfulness and Diversity Metrics Correlate with Human Judgements} We
estimate the extent to which automatic metrics of faithfulness and
diversity correlate with human ratings (using Spearman's rank
correlation coefficient) in
Table~\ref{table:xsum-correlation-results}. In line with previous
work \cite{maynez-etal-2020-faithfulness,kryscinski-etal-2019-neural},
we find that entailment scores are best correlated with faithfulness
(moderate, $0.40 \leq r \leq 0.59$). Like Self-BLUE, Self-Entailment
and Self-BERTScore are also strongly correlated with diversity
ratings. Compared to other metrics which capture a single
dimension, \ednascore is positively correlated with both dimensions of
diversity \emph{and} faithfulness.

Finally, in Figure~\ref{fig:score_diff_weights}, we plot faithfulness and diversity scores for different decoding strategies with varying temperatures and nucleus probabilities. We find that summaries
sampled with Composition(\frost$_{\hspace*{-.1cm}++}$)  are consistently more faithful than single-best Beam(\frost) summaries, but worse than summaries decoded with \mbox{Beam(\frost$_{\hspace*{-.1cm}++}$)}. Summaries sampled with \mbox{Composition(\frost$_{\hspace*{-.1cm}++}$)}  achieve the best \ednascore score (with $p=0.95$) amongst all diverse decoding strategies.

\section{Question Generation}
\label{sec:qgen}

\subsection{Dataset and Metrics} Question generation is often
conceptualized as the task of generating a question from a
passage-answer pair \cite{nqg2017}.  We experiment on
SQuAD \cite{rajpurkar-etal-2016-squad} and use the split
of \newcite{nqg2017} consisting of 86,635, 8,965, and 8,964
source-target pairs for training, validation, and testing,
respectively.\footnote{We also experimented with the split
of \newcite{du-etal-2017-learning}. Results can be found in the Appendix
(Table~\ref{table:qgen-results-dusplit}).}
We follow \newcite{cho-etal-2019-mixture} and report  \mbox{BLEU-4} (Top-1,
the single-best accuracy), Oracle (Top-5, the best accuracy among the
5-sampled hypotheses), and Self-BLEU (as defined
in Section~\ref{sec:singledocsum}).

\begin{table}[t]
 \begin{small}
  \begin{tabular}{@{\hspace*{-.2cm}}l@{~}l@{~}|@{~}c@{~}|@{~} c @{~}|@{~}c@{}} 
    \toprule
    & \multicolumn{1}{c@{~}|@{~}}{Models} & \multicolumn{1}{c}{T1} & T5 & S-BL \\ \midrule
    \multirow{4}{*}{\rotatebox[origin=c]{90}{{Single}}} & 
    NQG++ \cite{nqg2017} & 13.27 & --- & --- \\
    & MCP+GSA     \cite{zhao-etal-2018-paragraph}& 16.85 & --- & --- \\
    & \pegasus \citep{zhang2019pegasus} & \textbf{22.17} & --- & --- \\
    & \frost \citep{frost}& 21.04 & --- & --- \\ 
    
    \midrule
    \multirow{8}{*}{\rotatebox[origin=c]{90}{{Diverse}}} & 
    top-$k$ Sampling     \cite{fan-etal-2018-hierarchical} & 11.53 & 17.65 &  45.99 \\
    & Diverse Beam (\citeauthor{diversebeam}) & 13.38 & 18.30 & 74.80 \\ 
    & Mixture Decoder   \cite{pmlr-v97-shen19c} & 15.17 & 21.97 & 58.73 \\
    & Mixture Selector \cite{cho-etal-2019-mixture} & 15.67 & 22.45 &  59.82 \\
    & Mixture Selector \cite{wang-etal-2020-diversify} & 15.34 &  21.15 &  54.18 \\
    

    
    & Nucleus (\pegasus) & 12.05 & 24.72 & 30.64 \\
    & Nucleus (\frost)& 10.64 & 22.49 & \textbf{25.50} \\
    & Composition (\frost) & 17.16 & \textbf{27.04} & 61.68 \\
    & Composition (\frost$_{\hspace*{-.1cm}++}$) & \textbf{18.77} & 26.60 & 74.89 \\
    
    \bottomrule
  \end{tabular}
  \end{small}
  \caption{Comparison of different decoding techniques on question
    generation.  We report on BLEU-4 Top-1 accuracy (T1) and Top-5
    (T5), and Self-BLEU (S-BL).  Results for diverse decoding
    comparison models are taken from
    \newcite{wang-etal-2020-diversify}.  Best results in each block
    are bold-faced.}
  \label{table:qgen-results}
\end{table}

\subsection{Results} 

For our question generation experiments we also compare models which
employ single-best decoding against models which adopt diverse
decoding techniques. The top block in Table~\ref{table:qgen-results}
presents results for NQG++ \cite{nqg2017}, a pointer generator-based
model, CP+GSA \cite{zhao-etal-2018-paragraph}, a model which combines a pointer
mechanism with a gated self-attention encoder, and fine-tuned \pegasus
and \frost models. The second block in the table contains several
diverse decoding approaches including top-$k$
sampling \cite{fan-etal-2018-hierarchical}, diverse beam
search \cite{diversebeam}, mixture decoding \cite{pmlr-v97-shen19c}
and mixture content
selection \cite{cho-etal-2019-mixture,wang-etal-2020-diversify}. We
compare these models against nucleus sampling with \pegasus
and \frost, and composition sampling with \frost.
    
As in our summarization experiments, we observe that composition
sampling is not as performance diminishing as nucleus sampling, in
terms BLEU. \frost achieves a BLEU of $21.04$ (top-1) in the
single-best decoding setting; in comparison, BLEU drops for nucleus
sampling by $10.40$ points (on average), and~$2.27$ points only for
composition sampling (\frost$_{\hspace*{-.1cm}++}$).  Nucleus sampled questions achieve the best
pairwise diversity scores (Self-BLEU of~$25.50$), but very low
BLEU \mbox{Top-1} score of $10.64$. Composition sampled questions are
less diverse then other methods, but outperform all baselines on Top-1
and Oracle metrics. Poor diversity (in terms of Self-BLEU) in
composition sampled questions can be attributed to two limitations:
(a) SQuAD questions are mostly extractive, and (b) questions are
generated conditioned on the passage \emph{and} the answer spans;
leaving limited scope for models to generate diverse questions. An
example in the Appendix (Figure~\ref{fig:qgen-predictions-all})
demonstrates the effectiveness of composition sampling in generating
accurate and diverse questions compared to other sampling
methods.\footnote{Detailed comparative error analysis
to \newcite{cho-etal-2019-mixture}
and \newcite{wang-etal-2020-diversify} was not possible as their
predictions are not publicly available.}


\section{Conclusion}

We proposed Composition Sampling, a simple yet effective decoding method for faithful and diverse conditional generation. 
Our method is straightforward to implement and does not require any
external system to augment the input during inference. Our experiments demonstrate that it is currently the best available decoding strategy
for generating diverse and meaningful output. 
We also
introduced Self-Entailment and Self-BERTScore, to automatically
compute semantic diversity in summaries, and, \ednascore, for jointly
measuring faithfulness and diversity.


\section*{Acknowledgements}

We thank the reviewers, the ARR action editor, and the senior area chair for their valuable feedback. We would like to thank Ryan McDonald, Ankur Parikh, and Slav Petrov for their insightful comments. Many thanks also to Ashwin Kakarla and his team for their help with the human evaluation.



\section*{Ethical Considerations}

The nature of text generation leads to multiple ethical considerations
when considering applications. The main failure mode is that the model
can learn to mimic target properties in the training data that are not
desirable.
\paragraph{Faithfulness and Factuality} Since models create new text,
there is the danger that they may neither be faithful to the source
material nor factual. This can be exacerbated when the data itself has
highly abstractive targets, which require the model to generate words
not seen in the source material during training. This often leads  the model to generate content inconsistent with the source material \cite{maynez-etal-2020-faithfulness,kryscinski-etal-2020-evaluating,gabriel-etal-2021-go}.
\paragraph{Trustworthy Data} If the data itself is not trustworthy (comes from suspect or malicious sources) the model  will naturally become untrustworthy as it will ultimately learn the language and topics of the training data. For instance, if the training data is about Obama birther conspiracies, and the model is asked to generate information about the early life of Obama, there is a risk that false claims will be predicted by the model.
\paragraph{Bias in Data} Similarly, biases in the data around gender, race, etc., risk being propagated in the model predictions, which is common for most NLP tasks. This is especially true when the models are trained from non-contemporary data that do not represent current norms and practices \cite{blodgett-etal-2020-language}.

 The above considerations are non-malicious, in that the
model is merely learning to behave as its underlying source
material. If users of such models are not aware of these issues and do
not account for them, e.g., with better data selection and evaluation, then the generated text can be damaging.

 Generation models can also be misused in malicious ways. These include generating fake news, spam, and other text meant to mislead large sections of the general population.

\bibliography{anthology,frost}
\bibliographystyle{acl_natbib}

\appendix

\begin{table*}[t!]
  \begin{center}{\small
  \begin{tabular}{l  l | c c c | c c c } 
    \toprule
 &   Models  & \multicolumn{3}{c|}{XSum} & \multicolumn{3}{c}{CNN/DailyMail} \\
  &   & R1 &  R2 &  RL & R1 & R2 & RL \\
    \midrule
&    RoBERTaShare \cite{rothe-etal-2020-leveraging}& 38.52& 16.12& 31.13 & 39.25& 18.09& 36.45 \\
&    MASS  \cite{mass_icml19}& 39.75& 17.24& 31.95 & 42.12& 19.50& 39.01 \\
&    BART  \cite{bart}& 45.14& 22.27& 37.25 & 44.16& 21.28& 40.90 \\
&    GSum \citep{dou2020gsum} & 45.40& 21.89& 36.67 & \textbf{45.94}& 22.32& 42.48 \\
&    UniLM   \cite{unilm_arxiv19} & ---& ---& --- & 43.33& 20.21& 40.51 \\
&    T5  \cite{t5}& ---& ---& --- & 43.52& 21.55& 40.69 \\
&    ProphetNet  \cite{qi-etal-2020-prophetnet} & ---& ---& --- & 44.20& 21.17& 41.30 \\
   &    CTRLsum \citep{he2020ctrlsum}& ---& ---& --- & 45.65& \textbf{22.35}& \textbf{42.50} \\
&     \textsc{Fame}  \citep{aralikatte-etal-2021-focus} & 45.31& 22.75& 37.46 & 42.95& 20.79 &39.90 \\
&    \pegasus \citep{zhang2019pegasus} & 47.56& 24.87& 39.40 & 44.05& 21.69& 40.98 \\
&    \frost \cite{frost} & \textbf{47.80}& \textbf{25.06}& \textbf{39.76} & 45.11& 22.11& 42.01 \\
\raisebox{1.5cm}[0pt]{\begin{sideways}{Single}\end{sideways}}  &   \frost$_{\hspace*{-.1cm}++}$ \cite{frost}& 44.94& 21.58& 37.20 & 45.08& 22.14& 41.99 \\
    \midrule
  &  Focus (\textsc{Fame}) & 42.76& 19.89& 34.97 & ---& ---& --- \\
  &  Nucleus (\pegasus) & 38.49& 16.57& 30.99 & 36.27& 15.10& 33.46 \\
  &  Nucleus (\frost) & 40.26& 17.83& 32.49 & 38.49& 15.71& 35.49  \\
   & Composition (\frost) & \textbf{45.12}& \textbf{22.24}& \textbf{36.98} & 41.76& 18.94& 38.69 \\
\raisebox{.3cm}[0pt]{\begin{sideways}{Diverse}\end{sideways}}      &  Composition (\frost$_{\hspace*{-.1cm}++}$) & 43.82& 20.35& 35.89 & \textbf{42.37}& \textbf{19.48}& \textbf{39.28} \\
    
    \bottomrule
  \end{tabular}}
  \end{center}
  \caption{Full set of \rouge results on XSum and CNN/DailyMail test
    sets comparing different decoding techniques and SOTA models. 
    Best results in each block are bold-faced.}
  \label{table:xsum-cnndm-rouge-appendix}
\end{table*}

\begin{table}[t!]
  \begin{center}{\small
  \begin{tabular}{p{0.05cm} p{0.05cm} l | ccc } 
    \toprule
    \multicolumn{3}{c|}{\multirow{2}{*}{{Models}}} & \multicolumn{3}{c}{With Reference} \\
    & & & R1 & R2 &RL \\
    \midrule 
    \multirow{9}{*}{\rotatebox[origin=c]{90}{{XSum}}} & \multirow{4}{*}{\rotatebox[origin=c]{90}{{Single}}} & 
    Focus (\textsc{Fame})  & 41.20&20.30&34.23 \\
    & & \pegasus &  49.49&28.43&40.69 \\
    & & \frost  & 49.12&28.35&40.90 \\
    & & \frost$_{\hspace*{-.1cm}++}$) & 41.15&19.66&33.75 \\
    \cline{2-6}
    & \multirow{5}{*}{\rotatebox[origin=c]{90}{{Diverse}}} & Focus (\textsc{Fame})  & 36.58&16.32&29.19 \\
    & & Nucleus (\pegasus) & 38.91&18.43&31.10 \\
    & & Nucleus (\frost) & 41.96&20.77&33.81\\
    & & Composition (\frost) & \textbf{45.88}&\textbf{23.74}&\textbf{36.95} \\
    & & Composition (\frost$_{\hspace*{-.1cm}++}$) & 41.81&19.61&33.87  \\
    \midrule 
    \midrule 
    \multirow{7}{*}{\rotatebox[origin=c]{90}{{CNN/DailyMail}}} & \multirow{3}{*}{\rotatebox[origin=c]{90}{{Single}}} & \pegasus & 38.50&15.04&36.09 \\
    & & \frost & 41.89&17.54&39.03 \\
    & & \frost$_{\hspace*{-0.1cm}++}$ & 41.82&17.96&38.87 \\
    \cline{2-6}
    & \multirow{4}{*}{\rotatebox[origin=c]{90}{{Diverse}}} & Nucleus (\pegasus) & 31.57&10.62&28.99 \\
    & & Nucleus (\frost) &  34.62&11.78&31.58 \\
    & & Composition (\frost) &  \textbf{37.89}&14.88&35.06 \\
    & & Composition (\frost$_{\hspace*{-.1cm}++}$) &  37.79&\textbf{15.07}&\textbf{35.07} \\
    \bottomrule
  \end{tabular}}
  \end{center}
  \caption{Full set of \rouge results on 50 documents sampled from
    XSum and CNN/DailyMail (see also
    Table~\ref{table:xsum-cnndm-all-results} in the main paper).}
  \label{table:xsum-cnndm-all-results-appendix}
\end{table}

\begin{figure*}[t!]
  \center{\small
  \setlength\tabcolsep{0.1cm}
    \begin{tabular}{ p{15.5cm}}
    \toprule 
    \textbf{\gold:} Walsall have signed defender \textcolor{orangered}{Luke} Leahy on a \textcolor{orangered}{two-year} contract from Scottish \textcolor{orangered}{Championship} side Falkirk.\\
    \midrule
    \textbf{Input:} Leahy, 24, scored 12 goals in 158 appearances with Falkirk, having joined the club from Rugby Town in 2012. The left-back made 38 appearances last season, helping the club finish second in the Scottish second tier before they lost to Dundee United in the play-offs. He joins Walsall on a free transfer after his contract expired and is the League One club's first summer signing. Find all the latest football transfers on our dedicated page. \\
    \midrule
    \multicolumn{1}{c}{{Single-best summaries}} \\
    \midrule
     \textbf{\fame} ($d \rightarrow t_d \rightarrow s$):  Walsall have signed Falkirk defender \textcolor{orangered}{Alex} Leahy on a \textcolor{orangered}{two-year} deal. \\
 \textbf{\pegasus} ($d \rightarrow s$): Walsall have signed defender \textcolor{orangered}{Paddy} Leahy from Scottish \textcolor{orangered}{Championship} side Falkirk on a \textcolor{orangered}{three-year} deal. \\
 \textbf{\frost} $(d \rightarrow c;s)$:\content Walsall $|$ Falkirk $|$ \textcolor{orangered}{Liam} Leahy $|$ \textcolor{orangered}{two} \summary Walsall have signed Falkirk defender \textcolor{orangered}{Liam} Leahy on a \textcolor{orangered}{two-year} deal. \\
 \textbf{\frost} ($d \rightarrow c_{\text{drop}};s$): \content Walsall $|$ Falkirk $|$ Leahy \summary Walsall have signed Falkirk defender Leahy on a free transfer. \\
    \midrule
    \multicolumn{1}{c}{Focus Sampling: \textsc{Fame}}  \\
    \midrule
 $\mathbf{s_1 \rightarrow}$ \textcolor{orangered}{Welsall} have signed defender \textcolor{orangered}{Adebayu '' Adebayu ''eahy} on a \textcolor{orangered}{two-year} deal following his departure from Scottish \textcolor{orangered}{Championship} club \textcolor{orangered}{Falkiri}. \\
 $\mathbf{s_2 \rightarrow}$ \textcolor{orangered}{Welsall} have signed defender \textcolor{orangered}{Adebayu '' Adebayu ''eahy} on a \textcolor{orangered}{two-year} deal from Scottish \textcolor{orangered}{Championship} club \textcolor{orangered}{Falkock}. \\
 $\mathbf{s_3 \rightarrow}$ \textcolor{orangered}{Welsall} have signed defender \textcolor{orangered}{Adebayu '' Adebayu ''eahy} on a \textcolor{orangered}{two-year} deal from Scottish \textcolor{orangered}{Championship} club \textcolor{orangered}{Falkock}. \\
 $\mathbf{s_4 \rightarrow}$ \textcolor{orangered}{Welsall} have signed defender \textcolor{orangered}{Adebayu Leahys} from Scottish \textcolor{orangered}{Championship} club \textcolor{orangered}{Falk Falkiri} for an undisclosed fee on a \textcolor{orangered}{three-year} deal. \\
 $\mathbf{s_5 \rightarrow}$ \textcolor{orangered}{Welsall} have signed defender \textcolor{orangered}{Adebayu '' Adebayu ''eahny} on a \textcolor{orangered}{two-year} deal following his departure from Scottish \textcolor{orangered}{Championship} club \textcolor{orangered}{Falkock}. \\
    \midrule
    \multicolumn{1}{c}{Nucleus Sampling: \pegasus} \\
    \midrule
 $\mathbf{s_1 \rightarrow}$ Walsall have signed defender \textcolor{orangered}{Adam} Leahy from fellow Scottish Championship side Falkirk on a two-year contract. \\
 $\mathbf{s_2 \rightarrow}$ Walsall have signed defender \textcolor{orangered}{Matt} Leahy on a \textcolor{orangered}{two-year} deal from Falkirk. \\ 
  $\mathbf{s_3 \rightarrow}$ Walsall have signed Falkirk \textcolor{orangered}{full}-back \textcolor{orangered}{Tyrone} Leahy for \textcolor{orangered}{an undisclosed fee.} \\
 $\mathbf{s_4 \rightarrow}$ Walsall have signed defender \textcolor{orangered}{Jason} Leahy from Scottish \textcolor{orangered}{Championship} club Falkirk. \\
 $\mathbf{s_5 \rightarrow}$ Walsall have signed \textcolor{orangered}{Driscoll} defender \textcolor{orangered}{Chris} Leahy for \textcolor{orangered}{an undisclosed fee} from Scottish \textcolor{orangered}{Championship} side Falkirk. \\
    \midrule
    \multicolumn{1}{c}{Nucleus Sampling: \frost}\\
    \midrule
 $\mathbf{c_1;s_1 \rightarrow}$ \content Walsall $|$ \textcolor{orangered}{Rory} Leahy $|$ Falkirk  \summary \textcolor{orangered}{dawned on} Walsall as they signed defender \textcolor{orangered}{Rory} Leahy on a \textcolor{orangered}{season-long loan} from Falkirk. \\
 $\mathbf{c_2;s_2 \rightarrow}$ \content Walsall $|$ Falkirk $|$ \textcolor{orangered}{Liam} Leahy \summary Walsall have signed Falkirk defender \textcolor{orangered}{Liam} Leahy. \\ 
 $\mathbf{c_3;s_3 \rightarrow}$ \content  Falkirk $|$ \textcolor{orangered}{Wade} Leahy $|$ Walsall  \summary  Former Falkirk defender \textcolor{orangered}{Wade} Leahy has joined Walsall for an \textcolor{orangered}{undisclosed fee}. \\ 
 $\mathbf{c_4;s_4 \rightarrow}$  \content Walsall $|$ \textcolor{orangered}{Todd} Leahy $|$ Scottish \textcolor{orangered}{Championship} $|$ Falkirk  \summary  Walsall have signed defender \textcolor{orangered}{Todd} Leahy from Scottish \textcolor{orangered}{Championship} side Falkirk. \\ 
 $\mathbf{c_5;s_5 \rightarrow}$  \content Walsall $|$ \textcolor{orangered}{Greg} Leahy $|$ Scottish \textcolor{orangered}{Championship} $|$ Falkirk $|$ \textcolor{orangered}{two}   \summary  Walsall have signed defender \textcolor{orangered}{Greg} Leahy from Scottish \textcolor{orangered}{Championship} side Falkirk on a \textcolor{orangered}{two-year contract}. \\ 
    \midrule
    \multicolumn{1}{c}{Composition Sampling: \frost} \\
    \midrule
 $\mathbf{c_1;s_1 \rightarrow}$  \content Walsall $|$ \textcolor{orangered}{Rory} Leahy $|$ Falkirk \summary Walsall have signed defender \textcolor{orangered}{Rory} Leahy from Falkirk. \\
 $\mathbf{c_2;s_2 \rightarrow}$  \content Walsall $|$ Falkirk $|$ \textcolor{orangered}{Liam} Leahy \summary Walsall have signed Falkirk defender \textcolor{orangered}{Liam} Leahy.  \\
 $\mathbf{c_3;s_3 \rightarrow}$  \content Falkirk $|$ \textcolor{orangered}{Wade} Leahy $|$ Walsall \summary Falkirk defender \textcolor{orangered}{Wade} Leahy has joined Walsall. \\
 $\mathbf{c_4;s_4 \rightarrow}$  \content Walsall $|$ \textcolor{orangered}{Todd} Leahy $|$ Scottish \textcolor{orangered}{Championship} $|$ Falkirk  \summary Walsall have signed defender \textcolor{orangered}{Todd} Leahy from Scottish \textcolor{orangered}{Championship} side Falkirk. \\
 $\mathbf{c_5;s_5 \rightarrow}$  \content Walsall $|$ \textcolor{orangered}{Greg} Leahy $|$ Scottish \textcolor{orangered}{Championship} $|$ Falkirk $|$ \textcolor{orangered}{two}   \summary  Walsall have signed defender \textcolor{orangered}{Greg} Leahy from Scottish Championship side Falkirk on a \textcolor{orangered}{two-year deal}. \\
    \midrule
    \multicolumn{1}{c}{Composition Sampling \frost$_{\hspace*{-.1cm}++}$}\\ 
    \midrule
 $\mathbf{c_1;s_1 \rightarrow}$  \content Walsall $|$ Leahy $|$ Falkirk \summary Walsall have signed defender Leahy from Falkirk. \\
 $\mathbf{c_2;s_2 \rightarrow}$  \content Walsall $|$ Falkirk $|$ Leahy \summary Walsall have signed Falkirk defender Leahy on a free transfer. \\
 $\mathbf{c_3;s_3 \rightarrow}$  \content Falkirk $|$ Leahy $|$ Walsall \summary Falkirk defender Leahy has joined Walsall on a free transfer. \\
 $\mathbf{c_4;s_4 \rightarrow}$  \content Walsall $|$ Leahy $|$ Scottish $|$ Falkirk \summary  Walsall have signed defender Leahy from Scottish side Falkirk. \\
 $\mathbf{c_5;s_5 \rightarrow}$  \content Walsall $|$ Leahy $|$ Scottish $|$ Falkirk \summary Walsall have signed defender Leahy from Scottish side Falkirk. \\
    
    \bottomrule
    \end{tabular}     
  }
  \caption{Example input article, its human written summary, and 
    model predictions for the XSum dataset. We highlight spans in
    \textcolor{orangered}{orange} that are not faithful to the
    input. We use $c*$ and $s*$ to denote different compositions and their corresponding summaries. }
  \label{fig:xsum-predictions-all}
\end{figure*}

\begin{figure*}[t!]
  \center{\small 
  \setlength\tabcolsep{0.1cm}
    \begin{tabular}{ p{15.5cm}}

\textbf{\normalsize Chelsea star Eden Hazard vs Arsenal playmaker Santi Cazorla: As duo
prepare to reach 100 Premier League games, who has excited our experts
the most since 2012? }\\\\

      Chelsea's Eden Hazard and Arsenal's Santi
    Cazorla are set to reach a Premier League milestone this weekend
    when they each make their 100th appearance. \\
    Both players have been
    hugely influential since they moved to London in the summer of
    2012, but who has been the most exciting import to watch? \\
    Here,
    Sportsmail's reporters choose the player they most enjoy seeing in
    action. \\

    Eden Hazard (L) and Santi Cazorla are both set to make
    their 100th Premier League appearance this weekend. \\
    \textbf{Lee Clayton.}\\ Cazorla has wonderful balance. So does Hazard. Cazorla
    scores important goals. So does Hazard. Cazorla is two-footed. So
    is Hazard. Cazorla dances past opponents. So does Hazard. \\
    So,
    while there is not a lot to choose between them and Hazard is
    likely to get the most picks in this article, I am going for
    Cazorla. It's a personal choice. He is a wonderful footballer. I
    have paid to watch them both (and I will pay to watch them both
    again), but the little Spanish magician edges it for me. \\
    VERDICT:  CAZORLA. \\
    Cazorla, pictured in action against Burnley, has been an
    influential part of Arsenal's midfield this season. \\

    \textbf{Ian Ladyman.} \\ I
    remember when Manchester City baulked at paying Hazard's wages
    when the Belgian was up for grabs in 2012. Back then City thought
    the young forward had a rather high opinion of his own worth for a
    player who was yet to play in a major European league. \\

    In the
    early days of his time at Chelsea, it looked as though City may
    have been right. He showed flashes of brilliance but also looked
    rather too easy to push off the ball. \\

    Roll forward to 2015,
    however, and the 24-year-old has developed in to one of the most
    important players in the Barclays Premier League. Brave, strong
    and ambitious, Hazard plays on the front foot and with only one
    thought in this mind. \\
    Rather like Cristiano Ronaldo, he has also
    developed in to the type of player ever defender hates, simply
    because he gets back up every time he is knocked to the ground. He
    would get in every team in the Premier League and is one of the
    reasons Chelsea will win the title this season. \\VERDICT:
    HAZARD. \\Hazard controls the ball under pressure from Stoke
    midfielder Stephen Ireland at Stamford Bridge. \\

    \textbf{Dominic King.} It
    has to be Hazard. I saw him play for Lille twice in the season
    before he joined Chelsea – once against St Etienne, the other was
    what proved to be his final appearance against Nancy. He scored
    two in the first match, a hat-trick the latter and played a
    different game to those around him.\\ He hasn't disappointed since
    arriving here and I love the nonchalance with which he takes a
    penalty, his low centre of gravity and the way he can bamboozle
    defenders. If there is such a thing as £32million bargain, it is
    Hazard. \\VERDICT: HAZARD. \\Hazard celebrates after scoring a fine
    individual goal in Chelsea's 3-2 win against Hull in March. \\

    \textbf{Nick    Harris.}\\

    Now this is a tricky one because while Eden Hazard will
    frequently embark on a dribble or dink in a pass that will make
    you nod in appreciation, he'll also miss a penalty and make you
    groan. Whereas the older Cazorla, less flashy but no less of a
    technical master, is to my mind more of a fulcrum, more important
    relatively to the sum of Arsenal's parts than Hazard is to
    Chelsea. \\ You'll gasp at Hazard but Cazorla's wow factor is
    richer. That's not to dismiss either: both are brilliant
    footballers, contributing goals, assists and flair. Any neutral
    would bite your hand off to have either playing in your
    team. \\
    Forced to pick though, it's Cazorla, for his consistency and
    crucially doing it in the biggest games. Exhibit A would be
    Manchester City 0 Arsenal 2 in January; goal, assist, all-round
    brilliance, against a big team, at an important time.\\ VERDICT:
    CAZORLA. \\Cazorla scores from the penalty spot in Arsena's 2-0 away
    win at Manchester City in January. \\

    \textbf{Riath Al-Samarrai.}\\ Eden Hazard
    for me. Cazorla is an utter delight, a little pinball of a man who
    is probably the most two-footed player I've seen. Put him in a
    tight space and then you see what makes him rare among the
    best. But Hazard is the top player in the Premier League, in my
    opinion. \\ This is the sixth of his eight seasons as a professional
    where he has reached double figures and yet he offers so much more
    than goals (36 in 99 in the Premier League for Chelsea). He can
    beat a man and, better still, you sense he likes doing
    it. \\ Technically, his passing and shooting are excellent and he
    also has a mind capable of sussing out the shapes and systems in
    front of him. That intelligence, more specifically. \\
    \end{tabular}     
  }
  \caption{Input CNN/DailyMail article for summaries presented in Figure~\ref{fig:intro-cnndm-predictions-article1}.}
  \label{fig:cnndm-article1}
\end{figure*}

\begin{figure*}[ht!]
  \center{\small 
  \setlength\tabcolsep{0.1cm}
    \begin{tabular}{ p{13.5cm}}
    \toprule 
    \multicolumn{1}{c}{Human Written Summary} \\\midrule
    Nick Clegg made the admission in a rare joint interview with his wife Miriam. Lib Dem said she decided against moving into `government mansion'. `Discussion's a rather grand word for Miriam basically saying no,' he joked. Miriam claims he has put `country above party' at `great personal cost'. Tonight: Spotlight Nick Clegg tonight (Thursday) on ITV at 7.30pm. \\
    \midrule
    \multicolumn{1}{c}{{Single-best summaries using Beam Search}} \\
    \midrule
    \textbf{\pegasus}: Lib Dem leader reveals balance of power lies
    with wife Miriam Gonzalez Durantez. Couple invited cameras into
    the kitchen of their £1.5million family home. They revealed why
    they decided to remain with their three sons in London. Mrs Clegg
    said: `It has been the best decision for all of us and if I may,
    also for Nick as well because you can see how politicians
    sometimes can get in a bubble and be completely distant from
    normal society'. Mr Clegg insisted he felt more full of `vim and
    vigour' than he did in the start of the coalition. Focus groups
    conducted by the programme found Mr Clegg was characterised as the
    `chihuahua in a handbag' of the government. He said: `If you know
    you disagree with someone, there's no tension. I suspect meetings
    between Tony Blair and Gordon Brown were so fractious because they
    kind of should have agreed but they
    didn't'. \\ \multicolumn{1}{c}{} \\
     \textbf{\frost}: \content Nick Clegg $|$ Miriam $|||$ Lib Dem $|$ Whitehall $|||$  $|||$ Couple $|$ ITV $|$ Tonight $|||$ Putney $|$ London \summary Nick Clegg reveals balance of power lies with his wife Miriam in joint interview. Lib Dem leader was given option of moving family into Whitehall mansion. But the move was vetoed by his high flying lawyer wife after last election. Couple invited the cameras into their kitchen for ITV's Tonight programme. They were seen drinking wine and cooking in the kitchen of their home in Putney, south west London. \\ \multicolumn{1}{c}{} \\
 \textbf{\frost$_{\hspace*{-.1cm}++}$}: \content Nick Clegg $|$ Miriam $|||$ Lib Dem $|$ Whitehall $|||$ $|||$ ITV $|$ Tonight $|||$ Putney $|$ London \summary Nick Clegg reveals balance of power lies with his wife Miriam in joint interview. Lib Dem leader was given option of moving family into Whitehall mansion. But the move was vetoed by his high flying lawyer wife. The couple invited the cameras into their kitchen for ITV's Tonight programme. They were seen drinking white wine and cooking in their home in Putney, London. \\
    \bottomrule
    \end{tabular}     
  }
  \caption{Human written summary and single-best summaries using beam search for the CNN/DailyMail article in Figure~\ref{fig:cnndm-article2}.}
  \label{fig:beam-predictions-article2}
\end{figure*}

\begin{figure*}[ht!]
  \center{\small 
  \setlength\tabcolsep{0.1cm}
    \begin{tabular}{ p{15.5cm}}
    \toprule 
    
    \multicolumn{1}{c}{Nucleus Sampling: \pegasus} \\
    \midrule
     $\mathbf{s_1 \rightarrow}$ Nick Clegg reveals
    \textcolor{orangered}{balance902} of power lies with his wife
    Miriam Gonzalez Durantez \textcolor{orangered}{irish}. Couple
    performed joint interview \textcolor{orangered}{to promote new
      series} on ITV News Tonight. Cleggs spoke of pressure of moving
    from their Putney home into a taxpayer-funded mansion.  \\\multicolumn{1}{c}{}\\
      $\mathbf{s_2 \rightarrow}$  Lib Dem leader and wife Miriam give TV interview to ITV's Tonight program. Co-hosts have been pictured drinking white wine and cooking paella. They explained why she vetoed family heading to Whitehall mansion. Husband quipped: 'It's a grand word for Miriam basically saying no'. \\\multicolumn{1}{c}{}\\
     $\mathbf{s_3 \rightarrow}$ Lib Dem leader admitted wife Miriam has the final say over family life. Couple chose not to move their three \textcolor{orangered}{Laundry} to Whitehall home \textcolor{orangered}{earlier this May}. \\\multicolumn{1}{c}{}\\
     $\mathbf{s_4 \rightarrow}$  Nick Clegg and his wife Miriam Gonzalez Durantez open up in TV interview. Lib Dem leader revealed \textcolor{orangered}{she Bloomberg-style `discussions'} in their home. Couple revealed they \textcolor{orangered}{opted not to stay with their sons} in their £1.5m house. \\\multicolumn{1}{c}{}\\
     $\mathbf{s_5 \rightarrow}$  Liberal Democrats leader revealed balance of power lies \textcolor{orangered}{30-plus metres away}. He brought cameras into family home due to Cameron and Miliband controversies. Lib Dem leader joked that wife Miriam vetoed their move to Whitehall. \\
    \midrule
    \multicolumn{1}{c}{Nucleus Sampling: \frost} \\
    \midrule
 $\mathbf{c_1;s_1 \rightarrow}$ \content Liberal Democrats $|$ Nick Clegg $|$ Miriam Gonzalez Durantez $|||$ Putney $|$ London $|||$ Cleggs $|||$ ITV $|||$ Couple \summary Liberal Democrats leader Nick Clegg reveals balance of power with wife Miriam Gonzalez Durantez in joint interview. They invited cameras into kitchen of £1.5million family home in Putney, south west London. Cleggs are seen trying white wine as they discuss family life and \textcolor{orangered}{girlfriends}. They were\textcolor{orangered}{Furness on ITV programme} and said they chose home to protect family. Couple say choosing home stopped them veering off from wider society `in a bubble' \\\multicolumn{1}{c}{}\\
 $\mathbf{c_2;s_2 \rightarrow}$ \content Lib Dem $|$ ITV $|$ Tonight $|$ Miriam Gonzalez Durantez $|||$ $|||$ Couple $|$ Putney $|$ London \summary Lib Dem leader appeared on ITV's Tonight programme with wife Miriam Gonzalez Durantez. He was given the option of moving his family into a grace-and-favour government mansion but was vetoed. Couple invite cameras into family home in Putney, south west London to talk about family life. \\\multicolumn{1}{c}{}\\
 $\mathbf{c_3;s_3 \rightarrow}$ \content Lib Dems $|$ Miriam $|||$ Couple $|$ ITV $|$ Tonight $|||$ Putney $|$ London $|||$ bestseller $|$ Miliband \summary Lib Dems leader revealed balance of power lies with wife Miriam. Couple invited cameras into kitchen of their home for ITV's Tonight programme.Asked why they kept the family home \textcolor{orangered}{Galore} in Putney, south west London. Documentary follows \textcolor{orangered}{millions-selling bestseller's rave} over Miliband'!! \\\multicolumn{1}{c}{}\\
 $\mathbf{c_4;s_4 \rightarrow}$ \content Clegg $|$ Putney $|||$ $|||$ $|||$ Lib Dem \summary Mrs Clegg explains why the family stayed in their £1.5million home in Putney \textcolor{orangered}{1929}. Comparing their relationship to that of a different marriage, she said: `We just stand together and it's not any more of a difficulty than what many others have to go through'. \textcolor{orangered}{Revealingly, suggests that although no longer planning a political showdown} they are closer than they have ever been. Senior Lib Dem says he has learned how to ignore populist attacks.  \\\multicolumn{1}{c}{}\\
 $\mathbf{c_5;s_5 \rightarrow}$ \content Lib Dem $|$ ITV $|$ Tonight $|||$ David Cameron $|$ Lib Demaceae $|||$ Nick $|$ Miriam Gonzalez Durantez \summary Lib Dem leader allowed ITV's Tonight news programme into kitchen of their home. David Cameron \textcolor{orangered}{elegant kitchens sometimes look like Lib Demaceae}. Nick's high flying wife Miriam Gonzalez Durantez reveals balance of power is with her.  \\ 
    \bottomrule
    \end{tabular}     
  }
  \caption{Diverse summaries with nucleus sampling for the
    CNN/DailyMail article in Figure~\ref{fig:cnndm-article2}. We
    highlight spans in \textcolor{orangered}{orange} that are not
    faithful to the input document. We use $c*$ and $s*$ to denote
    different compositions and their corresponding summaries.}
  \label{fig:nucleus-predictions-article2}
\end{figure*}

\begin{figure*}[t!]
  \center{\small 
  \setlength\tabcolsep{0.1cm}
    \begin{tabular}{ p{15.5cm}}
    \toprule 
    
    \multicolumn{1}{c}{Composition Sampling: \frost}  \\
    \midrule
     $\mathbf{c_1;s_1 \rightarrow}$  \content Liberal Democrats $|$ Nick Clegg $|$ Miriam Gonzalez Durantez $|||$ Putney $|$ London $|||$ Cleggs $|||$ ITV $|||$ Couple \summary Liberal Democrats leader Nick Clegg reveals balance of power lies with wife Miriam Gonzalez Durantez. The couple invited the cameras into the kitchen of their £1.5million family home in Putney, south west London. Cleggs were seen drinking white wine and cooking in the kitchen as they chatted about their family life. Focus groups conducted by the ITV programme found he was characterised as the 'chihuahua in a handbag' of the government. Couple believe that choosing to remain in their family home has stopped them from becoming cut off from the rest of the world. \\\multicolumn{1}{c}{}\\
     $\mathbf{c_2;s_2 \rightarrow}$  \content Lib Dem $|$ ITV $|$ Tonight $|$ Miriam Gonzalez Durantez $|||$ $|||$ Couple $|$ Putney $|$ London \summary Lib Dem leader appeared on ITV's Tonight programme with wife Miriam Gonzalez Durantez. He was given the option of moving his family into a grace-and-favour government mansion - but the move was vetoed by his wife. Couple invited the cameras into the kitchen of their £1.5million family home in Putney, south west London. \\\multicolumn{1}{c}{}\\
     $\mathbf{c_3;s_3 \rightarrow}$  \content Lib Dems $|$ Miriam $|||$ Couple $|$ ITV $|$ Tonight $|||$ Putney $|$ London $|||$ bestseller $|$ Miliband \summary Lib Dems leader reveals balance of power lies with wife Miriam in joint interview. Couple invited the cameras into their kitchen for ITV's Tonight programme. They were seen drinking wine and cooking in their £1.5million home in Putney, south west London. Interview comes after bestseller's row over Miliband's small kitchen. \\\multicolumn{1}{c}{}\\
     $\mathbf{c_4;s_4 \rightarrow}$  \content Clegg $|$ Putney $|||$ $|||$ $|||$ Lib Dem \summary Mr Clegg and his wife invited the cameras into the kitchen of their Putney home. They were seen drinking wine and cooking as they chatted about their family life. The couple were asked why they decided to remain in their family home. Lib Dem leader was given the option of moving his family into a government mansion. \\\multicolumn{1}{c}{}\\
     $\mathbf{c_5;s_5 \rightarrow}$  \content Lib Dem $|$ ITV $|$ Tonight $|||$ David Cameron $|$ Lib \textcolor{orangered}{Demaceae} $|||$ Nick $|$ Miriam Gonzalez Durantez \summary Lib Dem leader reveals balance of power in joint interview with wife on ITV's Tonight. David Cameron's large country kitchen and Lib \textcolor{orangered}{Demaceae' small kitchen criticised}. Nick and Miriam Gonzalez Durantez reveal why they stayed at home.  \\
    \midrule
    \multicolumn{1}{c}{Composition Sampling:
      \frost$_{\hspace*{-.1cm}++}$} \\
    \midrule
 $\mathbf{c_1;s_1 \rightarrow}$  \content Liberal Democrats $|$ Nick Clegg $|$ Miriam Gonzalez Durantez $|||$ Putney $|$ London $|||$ Cleggs $|||$ ITV $|||$ \summary Liberal Democrats leader Nick Clegg revealed balance of power lies with wife Miriam Gonzalez Durantez. The couple invited the cameras into the kitchen of their £1.5million family home in Putney, south west London. The Cleggs were seen drinking white wine and cooking in the kitchen of their home. ITV programme found he was characterised as the 'chihuahua in a handbag' of the government. He also insisted he felt more full of 'vim and vigour' than he did in the start of the coalition. \\\multicolumn{1}{c}{}\\
 $\mathbf{c_2;s_2 \rightarrow}$  \content Lib Dem $|$ ITV $|$ Tonight $|$ Miriam Gonzalez Durantez $|||$ $|||$ Putney $|$ London \summary Lib Dem leader appeared on ITV's Tonight programme with wife Miriam Gonzalez Durantez. He was given the option of moving his family into a grace-and-favour government mansion - but the move was vetoed by his wife. The couple invited the cameras into the kitchen of their £1.5million family home in Putney, south west London. \\\multicolumn{1}{c}{}\\
 $\mathbf{c_3;s_3 \rightarrow}$ \content Lib $|$ Miriam $|||$ ITV $|$
    Tonight $|||$ Putney $|$ London $|||$ Miliband \summary Lib Dem
    leader reveals balance of power lies with wife Miriam in joint
    interview. The couple invited the cameras into their kitchen for
    ITV's Tonight programme. They were seen drinking wine and cooking
    in their £1.5million home in Putney, south west London. Comes
    after Miliband was widely mocked for posing with wife in his
    kitchen. \\\multicolumn{1}{c}{}\\
    $\mathbf{c_4;s_4 \rightarrow}$ \content Clegg $|$ Putney $|||$ $|||$ $|||$ Lib Dem \summary Mr Clegg and his wife invited the cameras into the kitchen of their Putney home. They were seen drinking wine and cooking as they chatted about their family life. The couple were asked why they decided to remain in their family home. Lib Dem leader was given the option of moving his family into a government mansion. \\\multicolumn{1}{c}{}\\
 $\mathbf{c_5;s_5 \rightarrow}$ \content Lib Dem $|$ ITV $|$ Tonight $|||$ David Cameron $|$ Lib $|||$ Nick $|$ Miriam Gonzalez Durantez  \summary Lib Dem leader reveals balance of power in joint interview with wife on ITV's Tonight. Comes after David Cameron invited cameras into Lib Dem leader's country kitchen. Nick and Miriam Gonzalez Durantez were seen drinking wine and cooking. \\
    \bottomrule
    \end{tabular}     
  }
  \caption{Diverse summaries with composition sampling for the
    CNN/DailyMail article in Figure~\ref{fig:cnndm-article2}. We
    highlight spans in \textcolor{orangered}{orange} that are not
    faithful to the input document. We use $c*$ and $s*$ to denote different compositions and their corresponding summaries. }
  \label{fig:composition-predictions-article2}
\end{figure*}

\begin{figure*}[t!]
  \center{\footnotesize 
  \setlength\tabcolsep{0.1cm}
    \begin{tabular}{ p{15.5cm}}
      \textbf{\normalsize Inside the Clegg kitchen: Over white wine
        and paella Nick reveals how Miriam put her foot down and
        refused to swap their family home for a grace-and-favour
        property} \\\\

      It is a conversation that will be familiar to couples across the
      country. What one spouse thinks is a 'discussion', the other
      understands they are being overruled. \\ In a joint interview
      with his high flying lawyer wife Miriam Gonzalez Durantez, Nick
      Clegg revealed the balance of power lies where many long
      suspected: with her. \\ After the last election, Mr Clegg was
      given the option of moving his family into a grace-and-favour
      government mansion - but the move was vetoed by his
      wife. \\ After controversies over David Cameron's large country
      kitchen and Ed Miliband's small second kitchen, the couple
      invited the cameras into the kitchen of their £1.5million family
      home in Putney, south west London for ITV's Tonight
      programme. Scroll down for video. \\ Home: In a revealing joint
      interview, Liberal Democrats leader Nick Clegg (pictured)
      admitted his wife Miriam (right) makes the big decisions in
      their household. \\ Mr Clegg is seen in the documentary drinking
      wine as his wife explains why she chose not to move her family
      into a government property. \\ They revealed why they decided to
      remain with their three sons Antonio, Alberto, and Miguel, in
      the family home instead of making the move to
      Whitehall. \\ Miriam, who uses her maiden name Gonzalez
      Durantez, told ITV News Political Editor Tom Bradby:\\
      'We had a lot of pressure at the time to go to one of the houses
      of the government. 'We discussed and thought the best thing
      would be for the children to stay here. \\ Revealingly, Mr Clegg
      quipped: 'Discussion's a rather grand word for Miriam basically
      saying no.' \\But he quickly added: 'You were so right, you were
      so right.' \\ However, the couple believe that choosing to
      remain in their family home has stopped them from becoming cut
      off from the rest of the world. \\ Mrs Clegg said: 'If you look
      at it with perspective it has been the best decision for all of
      us and if I may, also for Nick as well because you can see how
      politicians sometimes can get in a bubble and be completely
      distant from normal society and I think if you’re in your house
      in your neighbourhood, it’s much easier really.' \\
      The couple were asked why they decided to remain with their
      three sons Antonio, Alberto, and Miguel, in their £1.5million
      family home in Putney, south west London. \\ The couple believe
      that choosing to remain in their family home has stopped them
      from becoming cut off from the rest of the world. \\ Asked how
      they coped with the 'terrific kicking' given to her husband she
      said she didn't take it 'too seriously'. 'Just like any other
      marriage, we just stand together and it's not any more of a
      difficulty than what many others have to go through and you
      know. You should never take it too seriously.'\\
      And if he wanted five more years Mr Clegg said: 'Ten, 15, 20 why
      not! In for a penny, in for a pound.' \\
      He also insisted he felt more full of 'vim and vigour' than he
      did in the start of the coalition.\\ Focus groups conducted by
      the programme found Mr Clegg was characterised as the 'chihuahua
      in a handbag' of the government. When asked what kind of drink
      he was the participants settled on Babycham.\\ Asked how they
      coped with the 'terrific kicking' given to her husband, Mrs
      Clegg said she didn't take it 'too seriously' \\ The Cleggs were
      seen drinking white wine and cooking paella in the kitchen of
      their home as they chatted about their family life. \\ Honest:
      'Discussion's a rather grand word for Miriam basically saying
      no,' Mr Clegg (left) joked during the interview. \\ Ed Miliband
      was widely mocked after he posed with wife Justine in this
      picture, which turned out to be a second kitchen in his north
      London home used for 'tea and snacks' \\ David Cameron invited
      the cameras into his Oxfordshire home, where he revealed he did
      not plan to stand for a third term. \\ Mr Clegg sought to
      explain why his relations with the Prime Minister always seemed
      to be so cordial. He said: 'If you know you disagree with
      someone, there's no tension. I suspect meetings between Tony
      Blair and Gordon Brown were so fractious because they kind of
      should have agreed but they didn't. \\ 'When David Cameron and I
      sit in a meeting, as we do week in week out, we kind of know
      that our starting point is that we come from different vantage
      points...' \\ He claimed not to read all newspapers, and had
      learned how to ignore attacks form his opponents. \\ 'It sounds glib but I actually think you can't take it too seriously otherwise you spend all your time reacting to stuff and you just have to laugh at it because some of it is faintly silly.' \\Mrs Clegg added that their close bond as a family has protected from the political brickbats. \\'From my point of view if I spend my time thinking about whatever a specific person may has said, I don't have any time to do what I want to do.  \\
    \end{tabular}     
  }
  \caption{CNN/DailyMail input article for the summaries presented in Figures~\ref{fig:beam-predictions-article2}--\ref{fig:composition-predictions-article2}.}
  \label{fig:cnndm-article2}
\end{figure*}

\begin{figure*}[t!]
  \center{\small
  \setlength\tabcolsep{0.1cm}
    \begin{tabular}{ p{15.5cm}}
    \toprule 
    \textbf{\gold Question:} What does the Premier of Victoria need to lead in the Legislative Assembly?\\
    \midrule
    \textbf{Context with Answer (in boldface):} Answer: \textbf{most seats} <n> Context: The Premier of Victoria is the leader of the political party or coalition with the \textbf{most seats} in the Legislative Assembly. The Premier is the public face of government and, with cabinet, sets the legislative and political agenda. Cabinet consists of representatives elected to either house of parliament. It is responsible for managing areas of government that are not exclusively the Commonwealth's, by the Australian Constitution, such as education, health and law enforcement. The current Premier of Victoria is Daniel Andrews. \\
    \midrule
    \multicolumn{1}{c}{{Single-best summaries}} \\
    \midrule
 \textbf{\pegasus}: How many seats does the Premier of Victoria have in the Legislative Assembly? \\
 \textbf{\frost}: \content Premier $|$ Victoria $|$ Legislative Assembly \summary What does the Premier of Victoria have in the Legislative Assembly? \\
    \midrule
    \multicolumn{1}{c}{Nucleus Sampling: \pegasus} \\
    \midrule
 $\mathbf{s_1 \rightarrow}$ The Premier of Victoria would have how many seats in the Legislative Assembly? \\
 $\mathbf{s_2 \rightarrow}$ What is the \textcolor{orangered}{politician MP} expect to have in Legislative Assembly? \\
 $\mathbf{s_3 \rightarrow}$ Aside from being the leader of a political party or coalition, how is the \textcolor{orangered}{Premier of Victoria Geometry of the Legislative Assembly}? \\
 $\mathbf{s_4 \rightarrow}$ How many Legislative Assembly seats \textcolor{orangered}{is} the Premier of Victoria? \\
 $\mathbf{s_5 \rightarrow}$ \textcolor{orangered}{What are the Legislative Assembly seats?} \\
    \midrule
    \multicolumn{1}{c}{Nucleus Sampling: \frost}\\
    \midrule
 $\mathbf{c_1;s_1 \rightarrow}$ \content criteria $|$ Premier $|$ Victoria $|$ Coalition \summary What is \textcolor{orangered}{a Varied criteria} for a Premier of Victoria to possess in a Coalition? \\
 $\mathbf{c_2;s_2 \rightarrow}$ \content Premier $|$ Victoria $|$ leader $|$ party $|$ coalition $|$ Legislative Assembly \summary The Premier of Victoria isThe leader of the political party or coalition with to what in the Legislative Assembly? \\
 $\mathbf{c_3;s_3 \rightarrow}$  \content number $|$ Legislative Assembly $|$ seats $|$ Premier \summary What is the number of Legislative Assembly seats that the Premier holds? \\
 $\mathbf{c_4;s_4 \rightarrow}$ \content piece $|$ legislature $|$ leader $|$ party $|$ mixture $|$ members \summary \textcolor{orangered}{What piece of the legislature does the leader of the party have a mixture of members?} \\
 $\mathbf{c_5;s_5 \rightarrow}$  \content Premier $|$ Victoria $|$ Legislative Assembly \summary What does the Premier of Victoria have in the Legislative Assembly \\ 
    \midrule
    \multicolumn{1}{c}{Composition Sampling: \frost} \\
    \midrule
 $\mathbf{c_1;s_1 \rightarrow}$  \content Premier $|$ Victoria $|$ Legislative Assembly \summary What does the Premier of Victoria have in the Legislative Assembly? \\
 $\mathbf{c_2;s_2 \rightarrow}$ \content Premier $|$ party $|$ coalition $|$ Legislative Assembly \summary The Premier of the political party or coalition has what in the Legislative Assembly? \\
 $\mathbf{c_3;s_3 \rightarrow}$ \content Premier $|$ Victoria $|$ leader $|$ party $|$ Legislative Assembly \summary The Premier of Victoria is the leader of the political party with what in the Legislative Assembly? \\
 $\mathbf{c_4;s_4 \rightarrow}$ \content Premier $|$ Victoria $|$ party $|$ coalition \summary What does the Premier of Victoria have in his political party or coalition? \\
 $\mathbf{c_5;s_5 \rightarrow}$ \content Premier $|$ Victoria $|$ leader $|$ party $|$ coalition $|$ Legislative Assembly \summary The Premier of Victoria is the leader of the political party or coalition with what in the Legislative Assembly? \\
    
    \bottomrule
    \end{tabular}     
  }
  \caption{Example input passage with answer in boldface,  human
    written question, and  model predictions including diverse
    questions for the SQuAD Question Generation dataset. We highlight
    spans in \textcolor{orangered}{orange} that are not accurate with
    respect to the input context. We use $c*$ and $s*$ to denote
    different compositions and their corresponding questions.}
  \label{fig:qgen-predictions-all}
\end{figure*}

\begin{table}[t!]
  \begin{center}{\small
  \begin{tabular}{ l | c c c } 
    \toprule
    \multicolumn{1}{c|}{Models}     & R1 & R2 &RL \\
     \midrule
    \multicolumn{4}{c}{{Single-best with Beam Search}} \\
    \pegasus & 47.52 & 18.72 & 24.91  \\
    \frost & 43.12 & 16.93 & 22.49 \\
    \midrule
    \multicolumn{4}{c}{{Diverse Decoding, Average of five runs}} \\
    Nucleus (\frost) & 39.50 & 12.94 &19.50 \\
    Composition (\frost) & 42.47& 15.43 & 21.43 \\
    Composition (\frost$_{\hspace*{-.1cm}++}$) & 42.37 & 15.78 & 21.90 \\
    \midrule
    \multicolumn{4}{c}{{Diverse Decoding, Best of five runs}} \\
    Nucleus (\frost) & 44.40 & 16.86 & 23.03 \\
    Composition (\frost) & 46.98 & 19.34 & 24.96 \\
    Composition  (\frost$_{\hspace*{-.1cm}++}$) & 46.71 & 19.55 & 25.36 \\
    
    \bottomrule
  \end{tabular}}
  \end{center}
  \caption{\rouge results on the {Multi-News}
    \cite{fabbri-etal-2019-multi} multi-document summarization test
    set comparing different decoding techniques. The dataset contains
    56K articles in total paired with multi-line human-written summaries from the site \url{newser.com}. }
  \label{table:multidoc-results}
\end{table}

\begin{table}[t!]
  \begin{center}{\small
  \begin{tabular}{ l | c | c | c} 
    \toprule
    \multirow{2}{*}{Models} & BLEU-4 & Oracle & Pairwise \\
    & Top-1 & Top-5 & S-BLEU \\ \midrule
    \multicolumn{4}{c}{{Single-best with Beam Search}} \\
    \pegasus & \textbf{21.52} & --- & --- \\
    \frost & 19.98 & --- & --- \\ 

    \midrule
    \multicolumn{4}{c}{{Diverse Decoding}} \\

    Nucleus (\pegasus) & 12.60 & 24.45 & 31.23 \\
    Nucleus (\frost) & 10.98 & 22.61 & \textbf{26.36} \\
    Composition (\frost) & 16.62 & \textbf{26.07} & 62.47\\
    Composition (\frost$_{\hspace*{-.1cm}++}$) & \textbf{17.28} & 25.03 & 75.81 \\
    \bottomrule
  \end{tabular}}
  \end{center}
  \vspace{-0.3cm}
  \caption{We also experimented with the split of \newcite{du-etal-2017-learning} for SQuAD \cite{rajpurkar-etal-2016-squad}  question generation, consisting of 70,484, 10,570, and 11,877 examples for training, validation, and testing, respectively. Best results in each block are bold-faced.}
  \label{table:qgen-results-dusplit}
  \vspace{-0.5cm}
\end{table}

\end{document}